\documentclass[10pt, conference, compsocconf]{IEEEtran}

\usepackage{amsmath,array,graphicx}
\usepackage{graphics} % for pdf, bitmapped graphics files
\graphicspath{{images/},{images/sim_samples/},{images/real_samples/},{images/results/}}                          
\usepackage[]{subfig}
\usepackage{color}
\usepackage{soul}
\usepackage{multirow}
\usepackage{textcomp}
\usepackage[table]{xcolor}

\usepackage[hyphens]{url}
\usepackage[hidelinks]{hyperref}
\hypersetup{breaklinks=true}
\definecolor{y}{RGB}{255,255,0}
\definecolor{yo}{RGB}{254,191,0}
\definecolor{o}{RGB}{254,143,0}
\definecolor{or}{RGB}{254,106,0}
\definecolor{r}{RGB}{255,0,0}
\definecolor{rv}{RGB}{208,0,218}
\definecolor{v}{RGB}{174,0,255}
\definecolor{vb}{RGB}{140,0,255}
\definecolor{b}{RGB}{0,0,255}
\definecolor{bg}{RGB}{0,181,184}
\definecolor{g}{RGB}{0,255,0}
\definecolor{gy}{RGB}{172,255,0}

\usepackage{lipsum}

\let\OLDthebibliography\thebibliography
\renewcommand\thebibliography[1]{
  \OLDthebibliography{#1}
  \setlength{\parskip}{0pt}
  \setlength{\itemsep}{0pt plus 0.3ex}
}

\begin{document}
%
% paper title
% can use linebreaks \\ within to get better formatting as desired
\title{The Effect of Color Space Selection on Detectability and Discriminability of Colored Objects}

% author names and affiliations
% use a multiple column layout for up to two different
% affiliations

\author{\IEEEauthorblockN{Amir Rasouli and John K. Tsotsos}
\IEEEauthorblockA{Department of Electrical Engineering and Computer Science\\
York University\\
Toronto, Canada\\
\{aras,tsotsos\}@eecs.yorku.ca}
}

% make the title area
\maketitle

\begin{abstract}
In this paper, we investigate the effect of color space selection on detectability and discriminability of colored objects under various conditions. 20 color spaces from the literature are evaluated on a large dataset of simulated and real images. We measure the suitability of color spaces from two different perspectives: detectability and discriminability of various color groups. 

Through experimental evaluation, we found that there is no single optimal color space suitable for all color groups. The color spaces have different levels of sensitivity to different color groups and they are useful depending on the color of the sought object. Overall, the best results were achieved in both simulated and real images using color spaces \textit{C1C2C3}, \textit{UVW} and \textit{XYZ}.
   
In addition, using a simulated environment, we show a practical application of color space selection in the context of top-down control in active visual search. The results indicate that on average color space \textit{C1C2C3} followed by \textit{HSI} and \textit{XYZ} achieve the best time in searching for objects of various colors. Here, the right choice of color space can improve time of search on average by 20\%. As part of our contribution we also introduce a large dataset of simulated 3D objects. 

\end{abstract}


\IEEEpeerreviewmaketitle

\section{Introduction}
The choice of color space is an important task in various computer vision applications such as image compression \cite{Moroney1995luv}, and annotation \cite{Saber1996yes}, object detection \cite{Kumar2015}  or object tracking \cite{Danelljan2014hsv,Liang2015opp}. However, it is hard to define a universal color space as it can be modeled in numerous ways, e.g. \textit{Luv}, \textit{Lab}, \textit{HSV}, etc. \cite{Stokman2005}. 

The computer vision community has proposed a large number of solutions for color space selection for different applications. In one of the early works, Ohta  \textit{et al.} \cite{Ohta1980} introduce a new color space, \textit{I1I2I3} and show that using this color space can result in improved object segmentation. In \cite{Meas-Yedid2004}, the authors use an automatic color space selection for biological image segmentation. They use the Liu and Borsotti segmentation evaluation method to determine what color space provides the best segmentation. Similar adaptive approaches also have been applied to applications such as sky/cloud \cite{Dev2014} or skin \cite{Gupta2016} segmentation.

Using the right color space is also important in object detection and recognition applications. Vezhnevets \textit{et al.} \cite{Vezhnevets2003} perform a comparative study on various color spaces for skin detection. The authors highlight that changes in color luminance have little effect on separating skin from non-skin regions. In the context of cast shadow detection, Benedek and Sziranyi \cite{Benedek2007} evaluate numerous color spaces such as \textit{HSV}, \textit{RGB} and \textit{Luv}, and show that using \textit{Luv} color space is the most efficient for  color based clustering of the pixels and foreground-background-shadow segmentation. Van de Sande \textit{et al.} \cite{VanDeSande2010} evaluate different color spaces in conjunction with SIFT features for object detection. They argue that depending on the nature of the dataset using different color spaces such as opponent axis or \textit{RGB} can result in the best performance. Scandaliaris \textit{et al.} \cite{Scandaliaris2007} combine three color spaces including \textit{C1C2C3}, opponent axis and \textit{RGB} to generate shadow invariant features to detect objects by finding their contours. Moreover, there are a number of works that investigate the suitability of color spaces for texture classification of objects such as textile \cite{Paschos2001} and tree \cite{Porebski2007} classification.

In robotics, color spaces are also studied for various applications. Song \textit{et al.} \cite{Song2010} use a genetic algorithm to generate a color space suitable for recognition of colored objects in the context of robotic soccer. They show that using the color space, \textit{rSc2}, the highest recognition rate can be achieved compared to spaces such as \textit{YUV} or \textit{HSI}. In a similar study \cite{Song2014}, the authors propose the \textit{uSb} color space based on an iterative feature selection procedure for recognition of colored objects for underwater robotics. Duan \textit{et al.} \cite{Duan2015} investigate the optimal color space for segmentation of aerial images captured by unmanned UAVs. The authors report that for Bayer true color images, the \textit{I1I2I3} color space is the optimal choice for the segmentation of buildings compared to \textit{YCbCr}, \textit{YIQ} or \textit{Lab}. 

In the context of robotic visual search, color features can be used to optimize the process of search \cite{Rasouli2014vs}. In this work the authors use the sought object's color in a top-down control manner to bias the search. If the sought object is outside detection range, its color features are used to guide the search. The similarity between detected colors and the color of the object is measured using a backprojection technique. If a similarity is observed, the importance of the corresponding regions is increased, therefore the robot searches those regions first. The authors show that using this method, the time of search can be significantly reduced. In \cite{Rasouli2014vs}, however, the authors only use normalized \textit{RGB} color space and the search is only performed for a single red and green toy. 

In this paper we first investigate the impact of color space selection on detection of objects with different colors (detectability) and then using a cluster scoring technique identify how well different color groups can be separated (discriminability) using each color space. We perform this evaluation for the 20 most common color spaces in the literature using both synthetic and real images. Finally, we use active visual search as an application to examine the role of color space selection in practice. 

\section{Color space and robustness}
\label{sec:robustness}
\subsection{Color space}

We evaluated 20 color spaces (see Table \ref{table_color_spaces}) including the RGB space.

\begin{table}[!hbtp]
\caption{The list of color spaces evaluated} 
\vspace*{-\baselineskip} 
\label{table_color_spaces}
\begin{center}
\resizebox{0.8\columnwidth}{!}{
\begin{tabular}{|c|c|}
\hline
\textbf{Space} & \textbf{Reference}\\
\hline
\textit{XYZ, I1I2I3, HSI, YIQ} & Guo and Lyu \cite{Guo2000hsi} \\
\hline
\textit{Lab, YCrCb, rg, HSV} & Danelljan \textit{et al.} \cite{Danelljan2014hsv}\\
\hline
\textit{C1C2C3} & Salvador \textit{et al.} \cite{salvador2004cast}\\
\hline
\textit{Opp, Nopp, Copp} & Liang \textit{et al.} \cite{Liang2015opp}\\
\hline
\textit{Luv, xyz} & Moroney and Fairchild \cite{Moroney1995luv}\\
\hline
\textit{YES} & Saber \textit{et al.} \cite{Saber1996yes}\\
\hline
\textit{CMY, YUV} & Tkalcic and Tasic \cite{Tkalcic2013yuv} \\ 
\hline
\textit{HSL} &  Weeks \textit{et al.} \cite{Weeks1995hsl} \\ 
\hline
\textit{UVW} & Ohta \textit{et al.} \cite{Ohta1980uvw}\\
\hline
\textit{xyY} & Lucchese and Mitra  \cite{Lucchese2000xyy}\\
\hline
\end{tabular}
}
\end{center}
\end{table}
\vspace*{-\baselineskip} 

As mentioned earlier, our objective is to evaluate the suitability of color spaces for various applications, in particular, detecting and distinguishing colored objects. In this sense, two factors have to be considered. First, the color should be represented in a way that it is easily detectable (\textit{Detectability}) under various conditions such as in the presence of shadow, illumination changes or various reflecting surfaces. Second, different color groups should be easily separable (\textit{Discriminability}) and not confused with one another. Measuring the \textit{Detectability} and \textit{Discriminability} of colors in different color spaces accounts for how robust a color representation can be.

\subsection{Measuring robustness}

\subsubsection{Detectability}

We measure the detectability of different colors using the histogram backprojection (BP) technique \cite{Swain1991}. The BP algorithm generates a probability map in which the pixel values refer to how likely is the presence of a given color in the image. The computation of the BP map is as follows. Let $h(C)$ be the histogram function which maps color space $C = (a_1,a_2,...,a_i)$ , where $a_i$ is the $i^{th}$ channel of $C$, to a bin of histogram $H(C)$ computed from the object\textquotesingle s template, $T_\Theta$. The backprojection of the object's color over an image $I$ is given by,  
\begin{align}
\forall x,y : b_{x,y} := h(I_{x,y,c})
\end{align}

\noindent where $b$ is the grayscale backprojection image.

Choosing the right bin size for histograms is vital in the BP algorithm. The larger the size of the bins, the more tolerant BP is to the illumination change. At the same time different colors will likely be detected together. To eliminate bias in our detection, we use histograms with bin sizes of 16, 32, 64 and 128 and average the results over all configurations.

We compare the detection results against the ground truth data and use 3 measures of performance, $Recall (R) = T_{pos}/{T_{pos}+F_{neg}} $, $Precision (P) = T_{pos}/{T_{pos}+F_{pos}}$ and $FMeasure = {2*R*P}/{R+P}$ where $T_{pos} , F_{neg} $, and  $F_{pos}$ stand for true positive, false negative and false positive respectively.

\subsubsection{Discriminability}

\begin{figure}[!hbtp]
\centering
\includegraphics[width=0.25\textwidth]{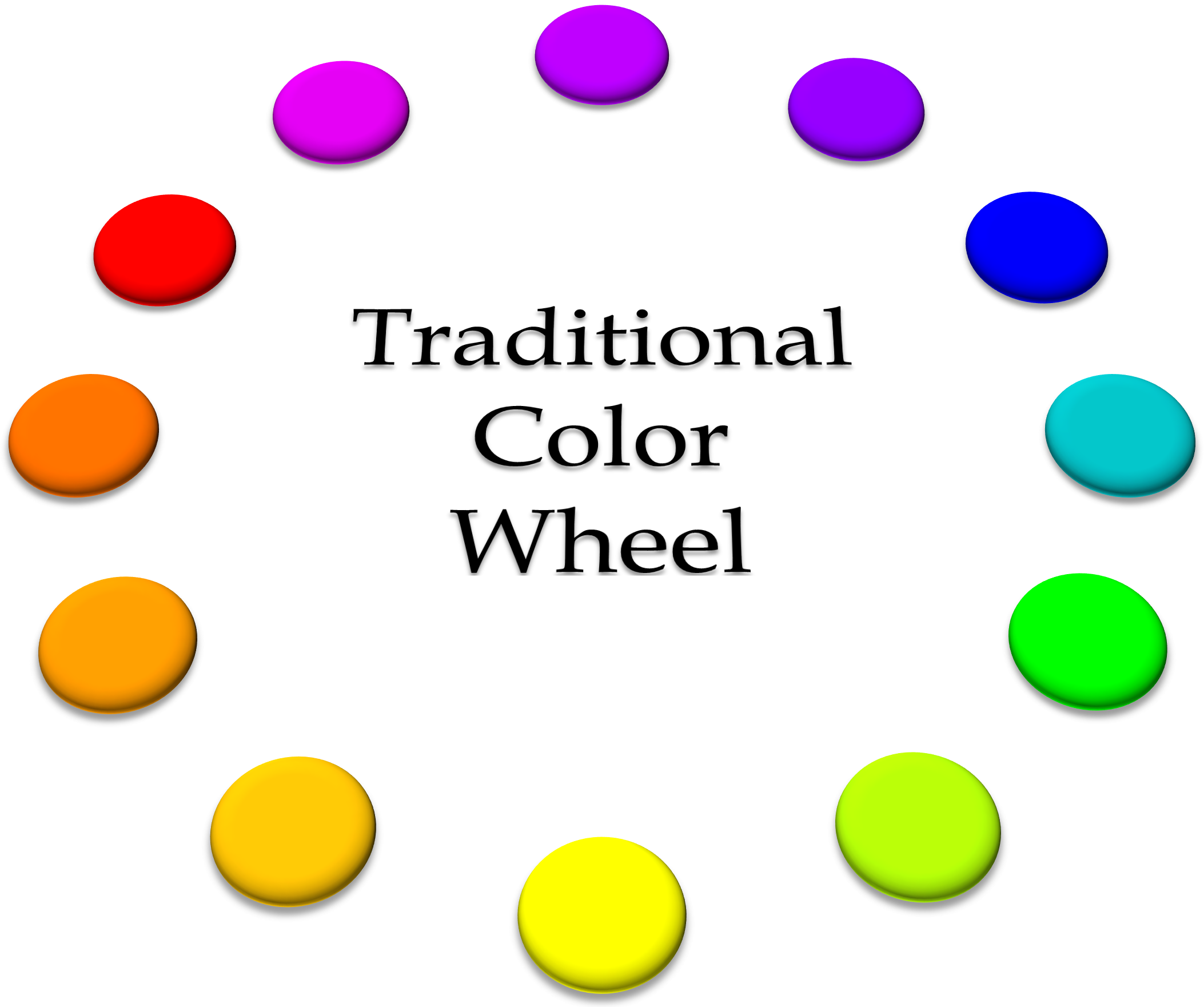} 
\caption{The traditional color wheel} 
\label{fig:color_wheel}
\end{figure}

One way to measure discriminability (separability) of colors in different spaces is by performing clustering and measuring how well the color data is grouped into different clusters. For this purpose we use a \textit{K-means} clustering technique using Expectation Maximization (EM) to find the maximum number of clusters that best represent the color distributions in each image. The number of clusters depends on the number of colors (out of 12 colors of the traditional color wheel, see Figure \ref{fig:color_wheel}) present in the input image.

We employ silhouette analysis \cite{rousseeuw1987silhouettes} to measure the consistency of each cluster against the ground truth. 
Let $i$ be a single pixel in the image and $clust_p$ be the cluster that $i$ belongs to. We define $a(i)$ as the average dissimilarity of pixel $i$ to all other pixels in cluster $clust_p$.
We define $d(i,CL)$ as the average dissimilarity of $i$ to all other pixels in  $CL = \{ clust_1,clust_2, ..., clust_j\}, j\neq p$ and $b(i) = \min d(i,CL)$. Based on these definitions, the silhouette measure $s(i)$ of pixel $i$ is given by, 
\begin{align}
s(i) = {b(i)-a(i)}/ \max(a(i)-b(i)), 
\end{align}
where $-1\leq s(i) \leq 1$. The value of 1 means pixel $i$ is well clustered whereas -1 implies that it does not belong to the allocated cluster.  

\section{Test Images}
\subsection{Synthetic images}
We setup a simulated environment to capture the robustness of the color spaces to various illumination conditions. The following setups were considered for our simulation.

\noindent \textit{Objects - } To model different surfaces we used objects with three different shapes namely, sphere, cylinder and cube (see Figure \ref{fig:objects})

\begin{figure}[!hbtp]
\centering
\includegraphics[width=0.20\textwidth]{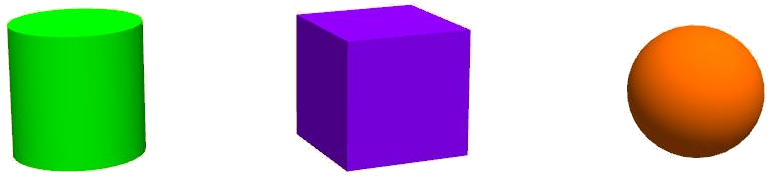} 
\caption{The sample shapes used in our experiments.} 
\label{fig:objects}
\end{figure} 

\noindent \textit{Colors - } We used 12 colors of the traditional color wheel depicted in Figure \ref{fig:color_wheel}. The traditional color wheel is chosen because it represents three basic color schemes including primary, secondary and tertiary colors.  

\begin{figure}[!hbtp]
\centering
\includegraphics[width=0.30\textwidth]{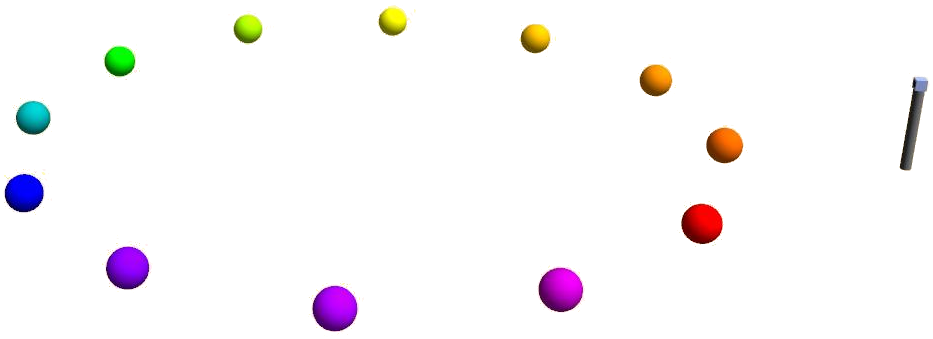} 
\caption{The circular configuration of objects. The rightmost object is the 3D model of the camera.} 
\label{fig:envconfig}
\end{figure}

\noindent \textit{Configuration - } The objects where placed on a circle with the radius of 2 meters from the center and equally distant from one another (see Figure \ref{fig:envconfig}). 

\noindent \textit{Camera - } We used a camera model same as the Zed stereo camera with $110^o$ FOV and 1280 $\times$ 1024 resolution placed on a rod with the height of 1 meter. The camera was rotated 12 times along a circle with radius of 3 meters and each time it was placed 1 meter away from the closest object while facing towards it. 

\begin{figure}[!hbtp]
\centering
\includegraphics[width=0.28\textwidth]{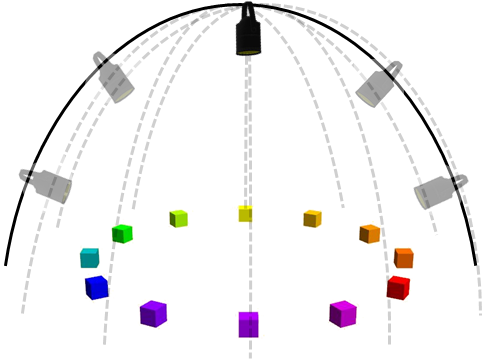} 
\caption{The rotation path of the point light source above the objects.} 
\vspace*{-\baselineskip} 
\label{fig:campose}
\end{figure}

\noindent \textit{Light - } We used two types of light sources, directional and point, placed 10 meters above the scene. To vary illumination conditions, the directional light source was rotated along the y-axis with an interval of $\pi/6$ forming a total of 12 orientations. The point light source also was rotated along an arc in the x-z plane by $\pi/6$ within an interval of $[-\pi/3, \pi/3]$. The arc also was rotated along z-axis 6 times equally within an interval of $[-\pi/2,\pi]$ (see Figure \ref{fig:campose}). 

\begin{figure}[!tbtp]
\captionsetup[subfigure]{labelformat=empty}
\centering
\subfloat[]{\includegraphics[width=0.12\textwidth]{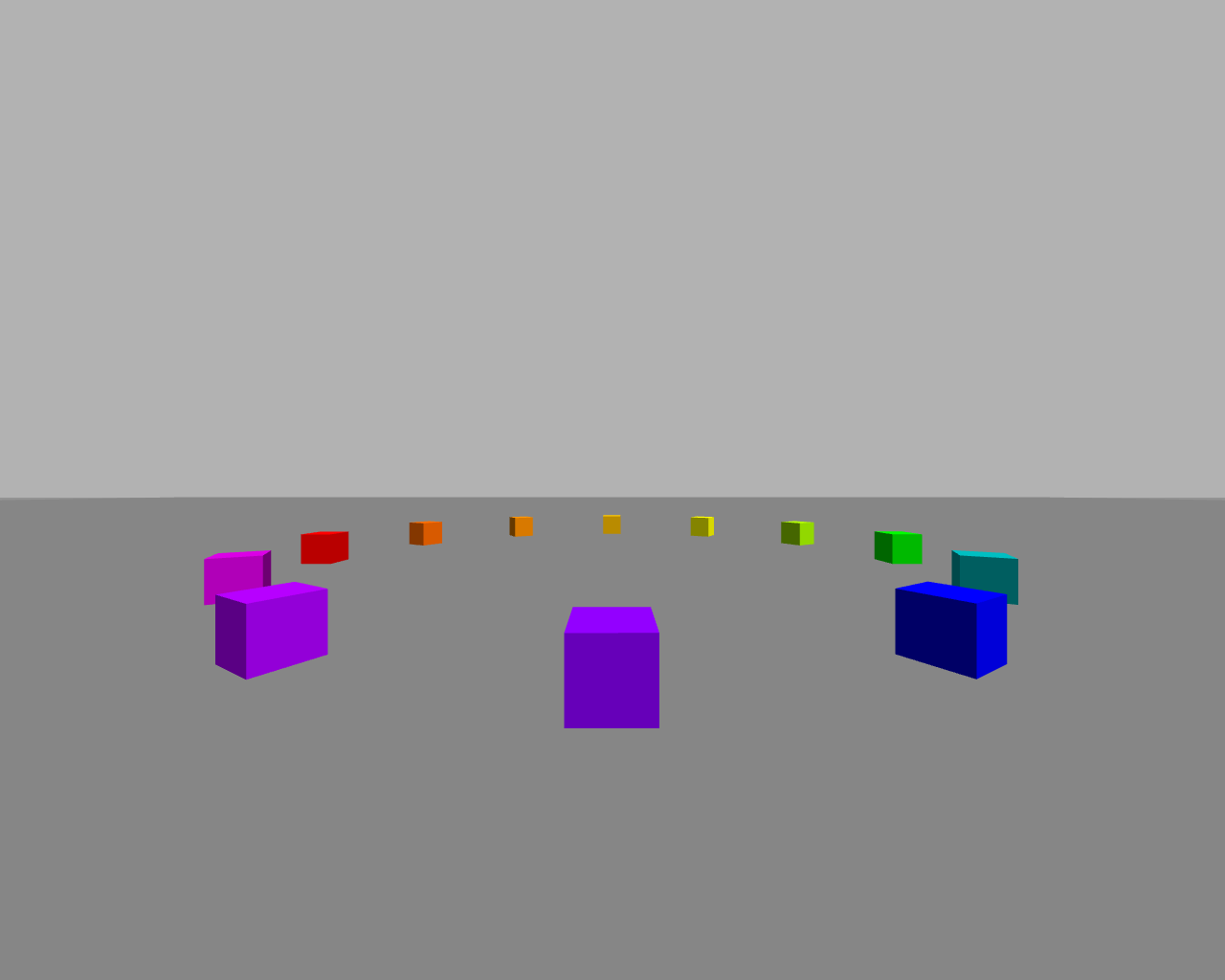}}
\hspace{0.2cm}
\subfloat[]{\includegraphics[width=0.12\textwidth]{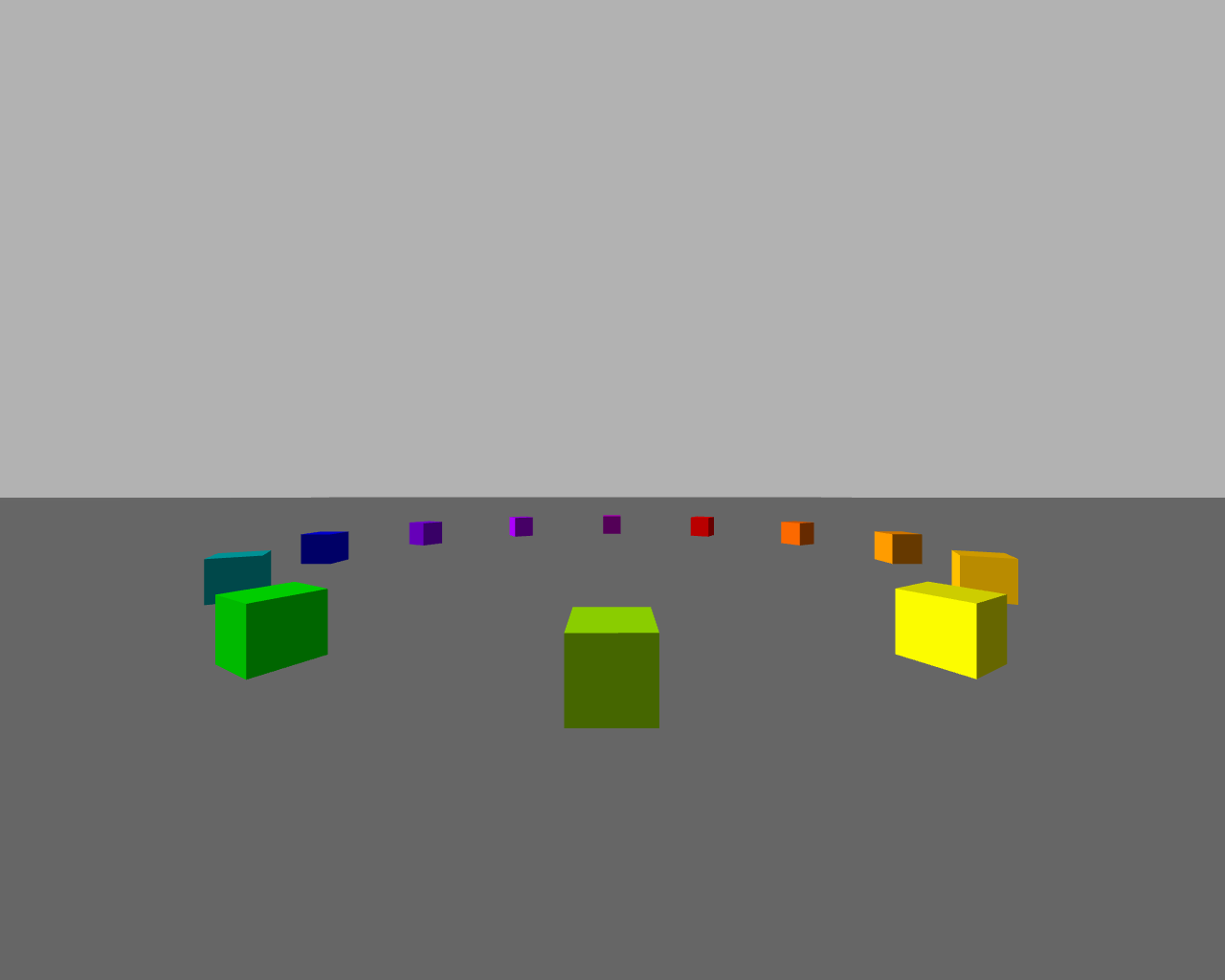}}
\hspace{0.2cm}
\subfloat[]{\includegraphics[width=0.12\textwidth]{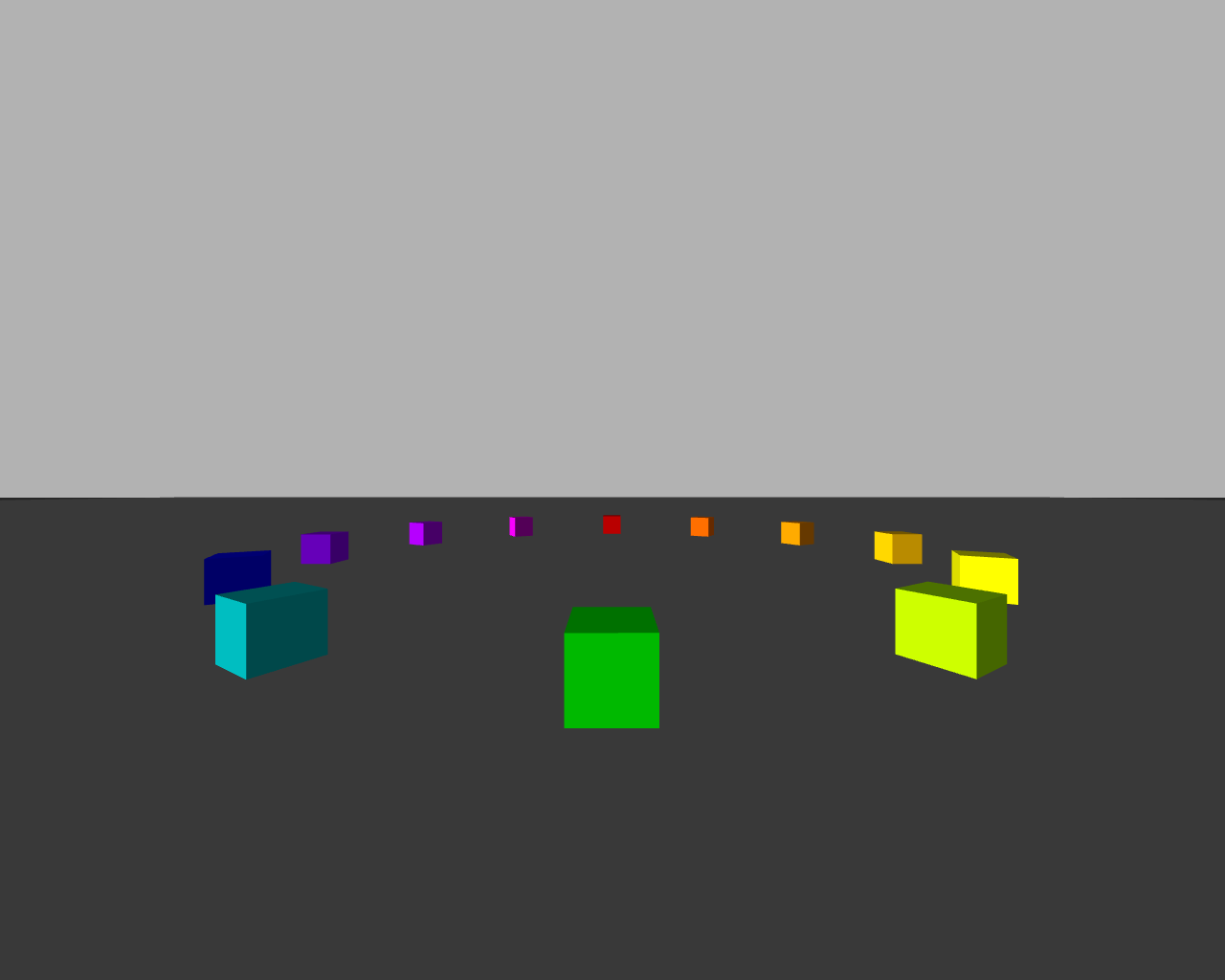}}\\
\vspace*{-\baselineskip} 
\subfloat[]{\includegraphics[width=0.12\textwidth]{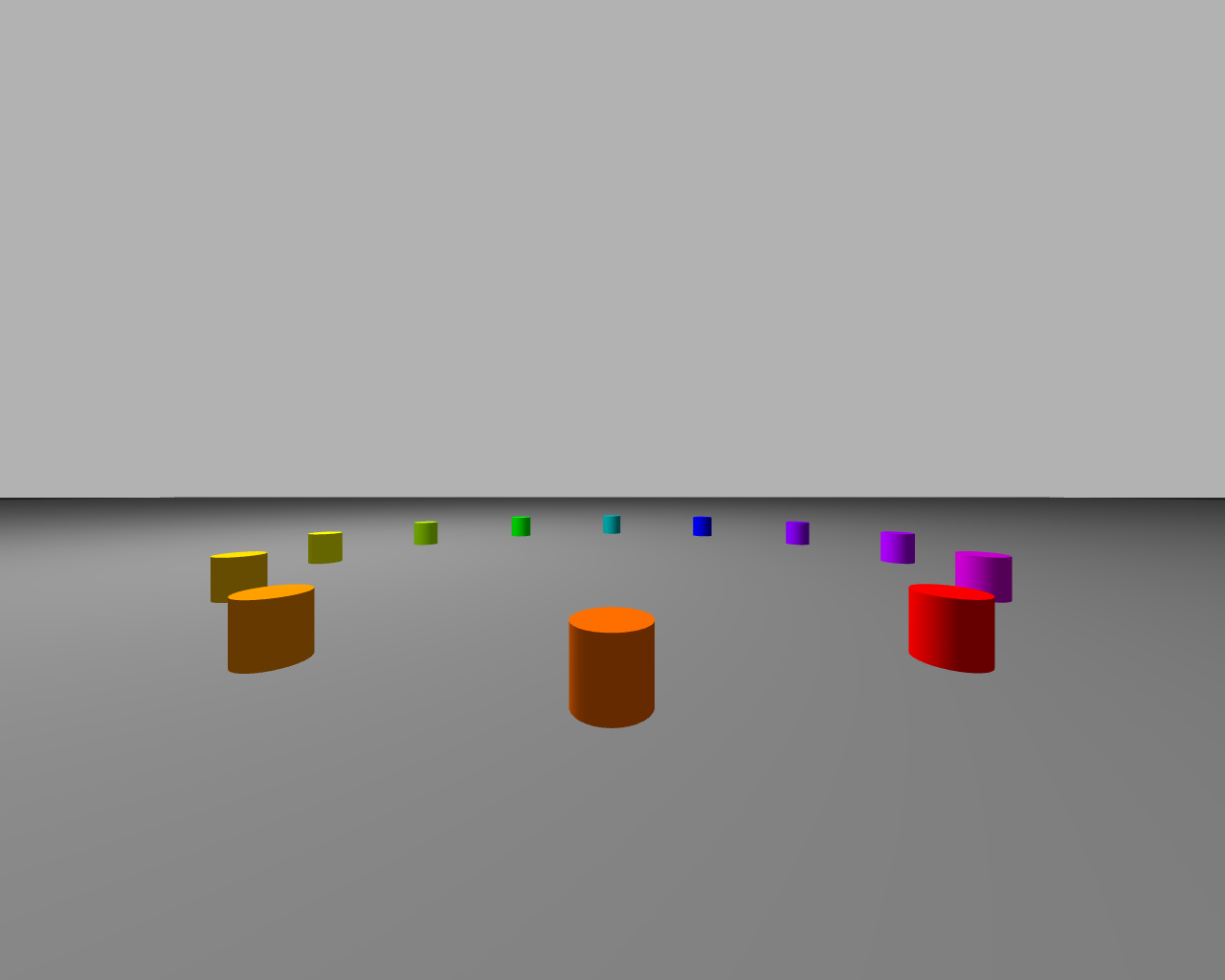}}
\hspace{0.2cm}
\subfloat[]{\includegraphics[width=0.12\textwidth]{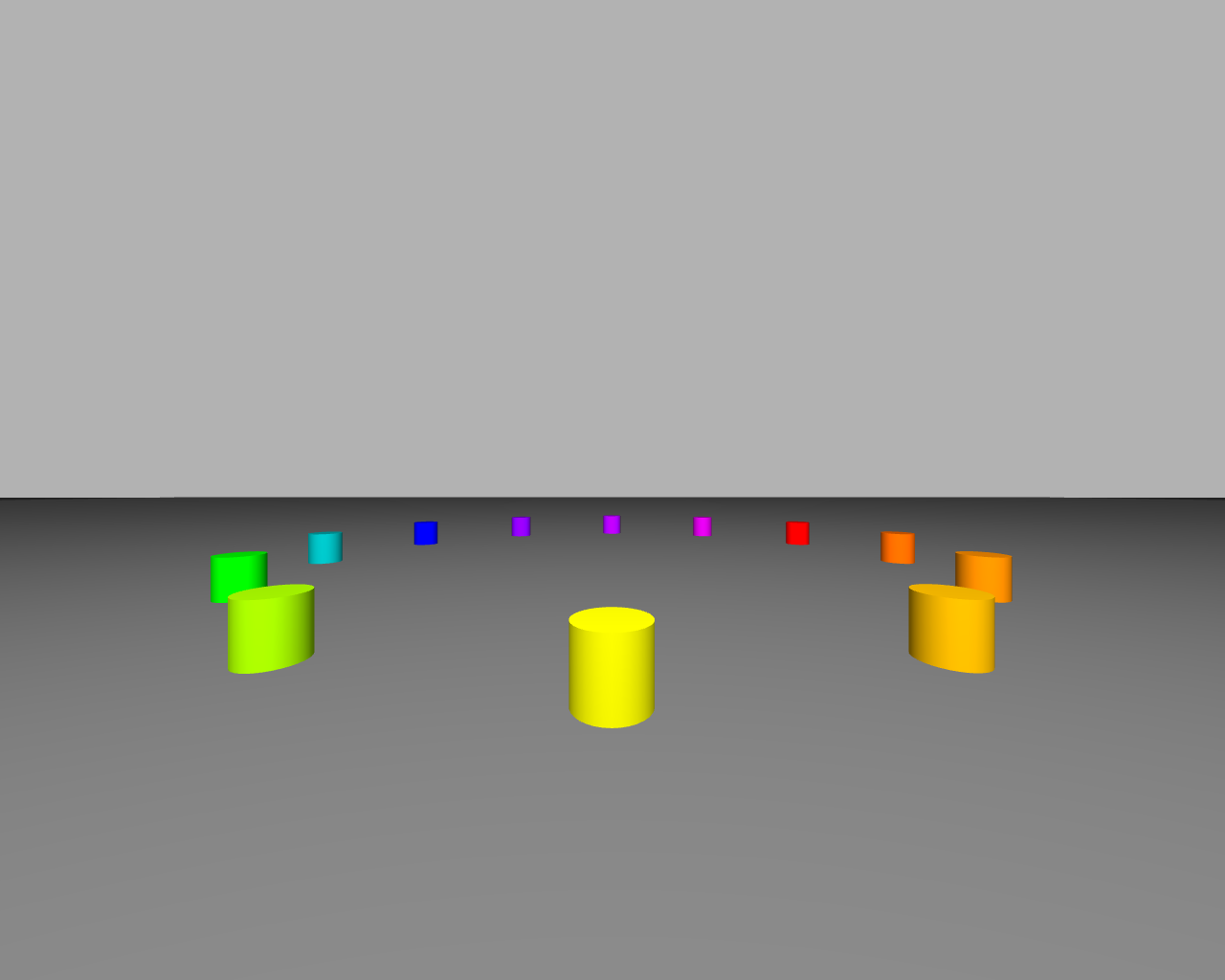}}
\hspace{0.2cm}
\subfloat[]{\includegraphics[width=0.12\textwidth]{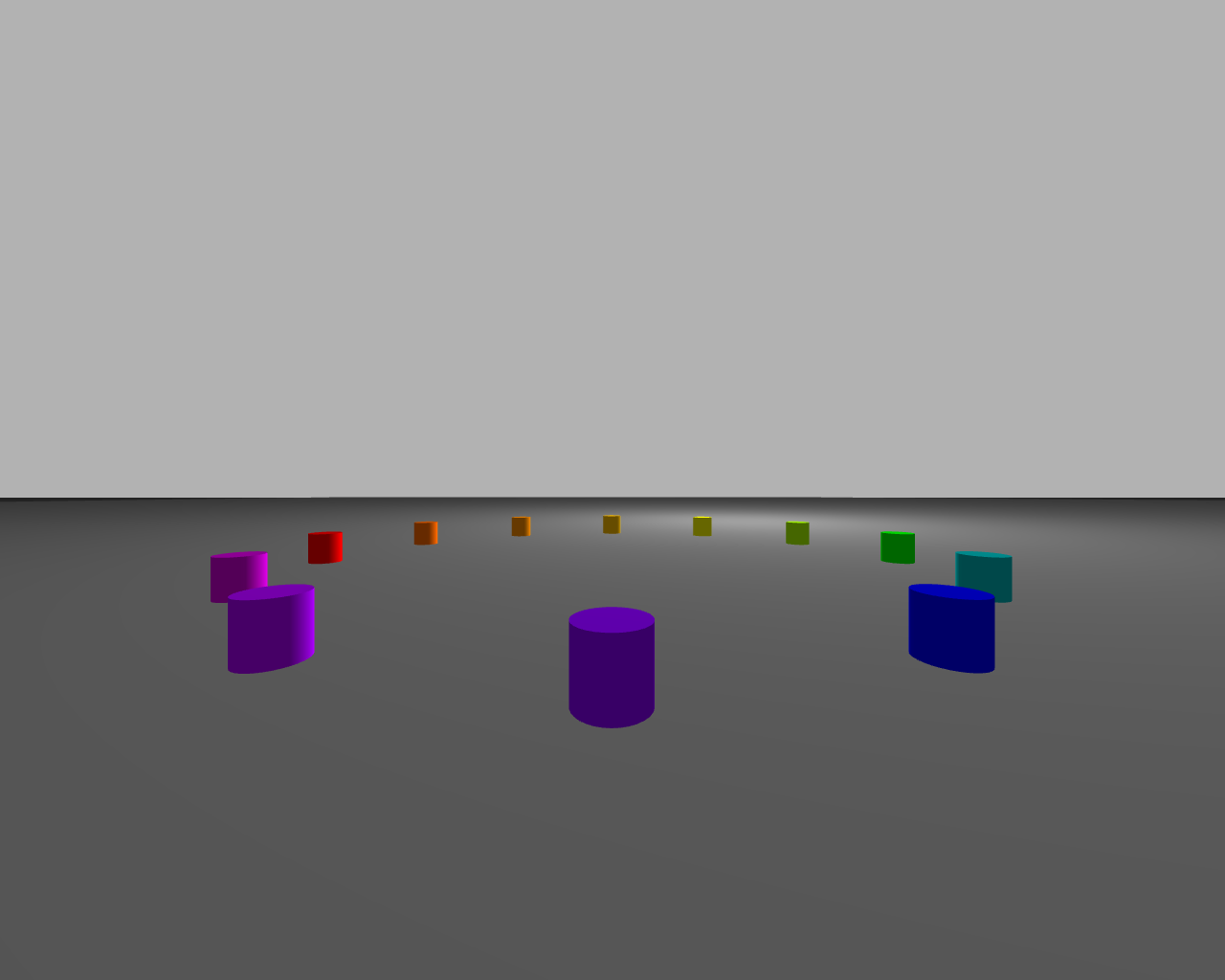}}\\\vspace*{-\baselineskip} 
\subfloat[]{\includegraphics[width=0.12\textwidth]{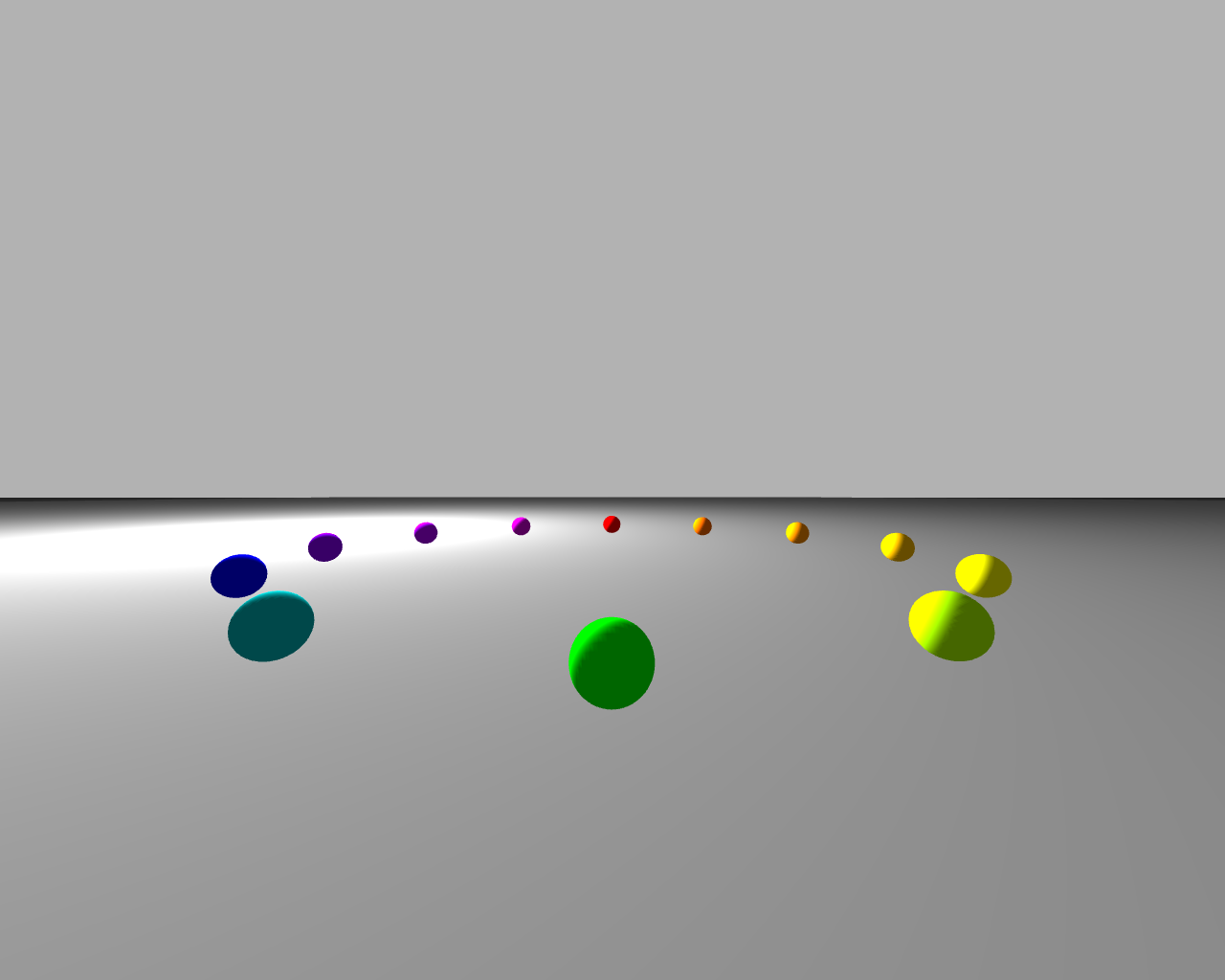}}
\hspace{0.2cm}
\subfloat[]{\includegraphics[width=0.12\textwidth]{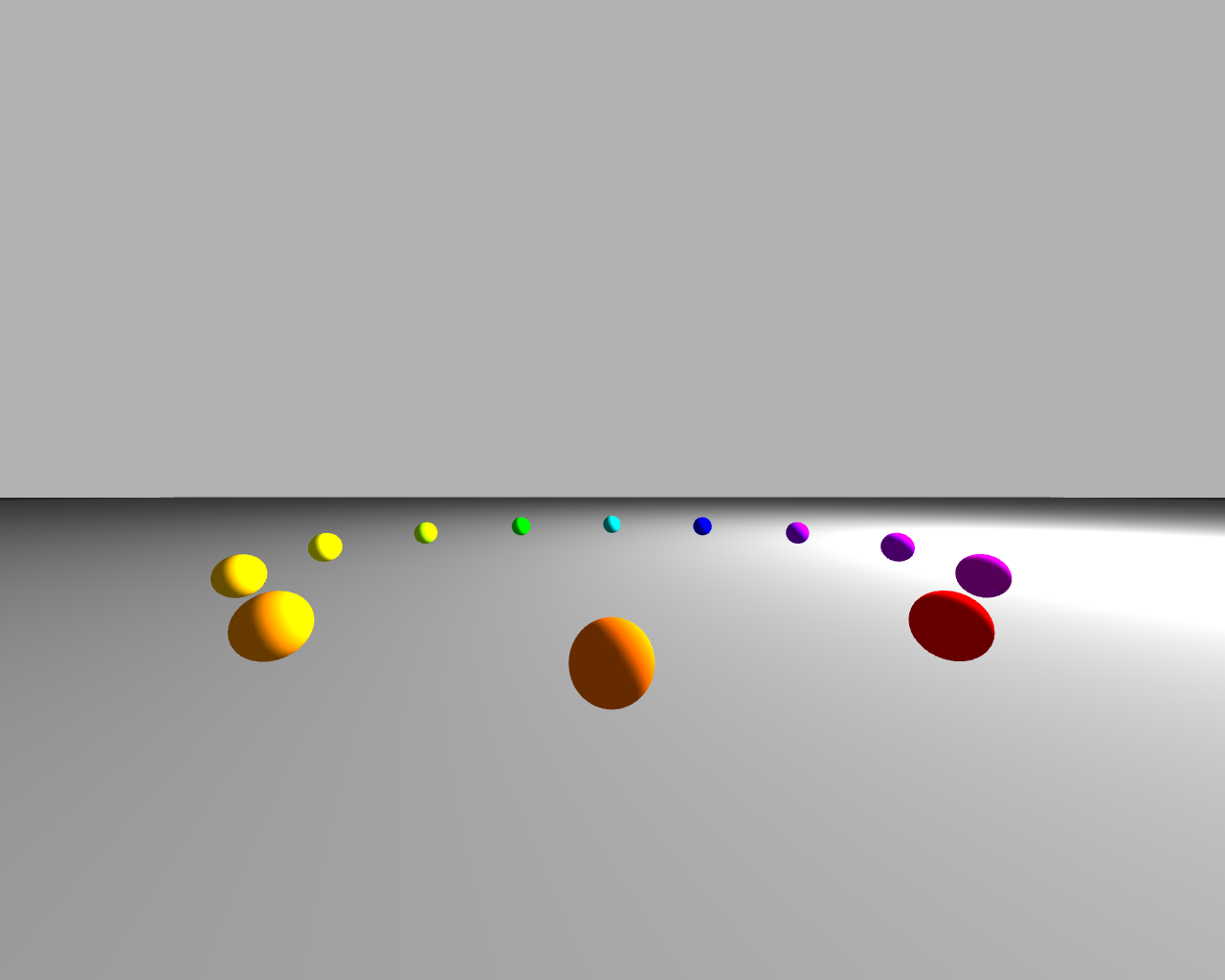}}
\hspace{0.2cm}
\subfloat[]{\includegraphics[width=0.12\textwidth]{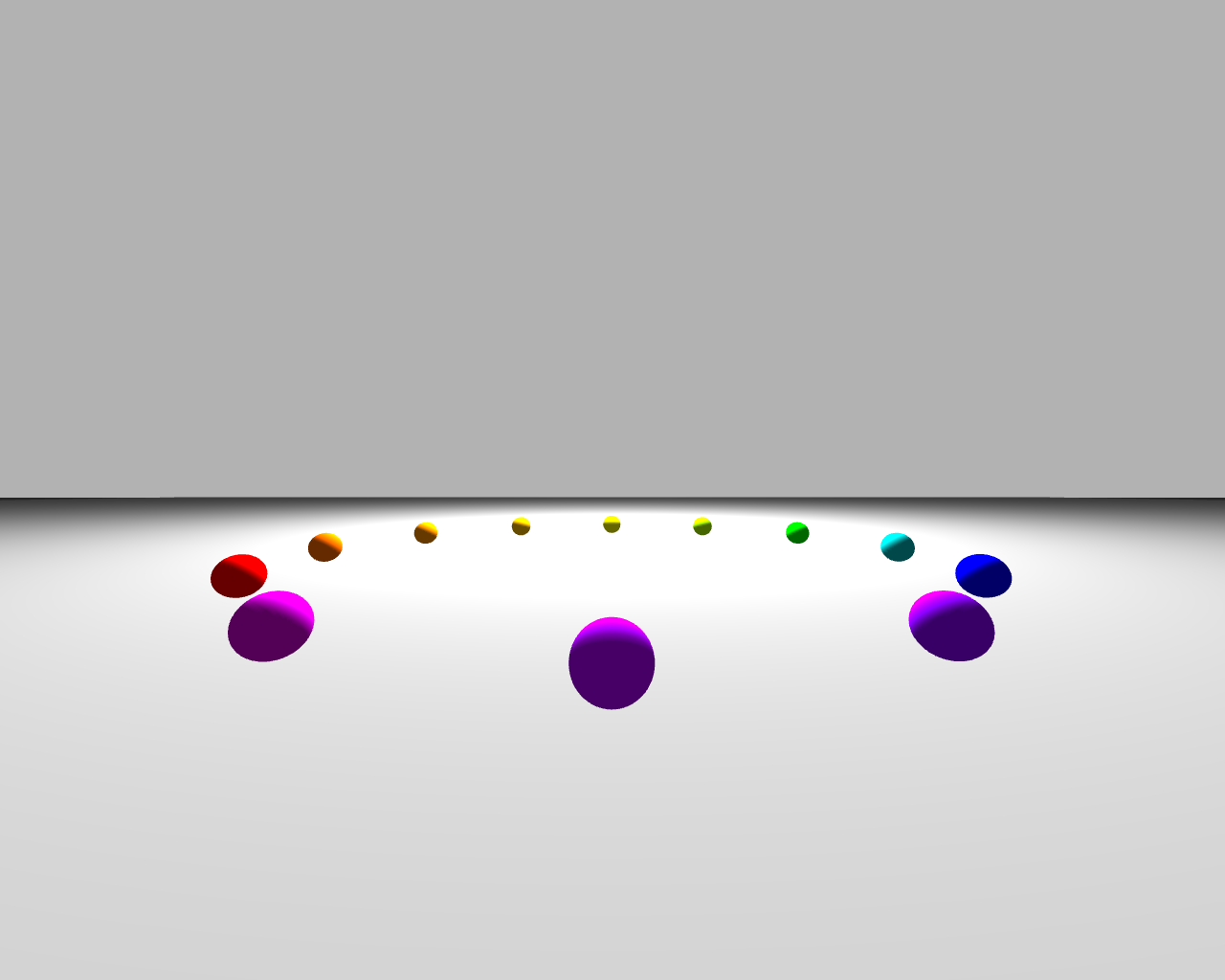}}
\vspace*{-\baselineskip} 
\caption{Synthetic image samples generated under various lighting conditions. From the top: directional light and point light with low and high intensity.}
\label{fig:sim_samples}
\end{figure}

Using the above setups a total of 4752 simulated images where generated. Figure \ref{fig:sim_samples} shows some sample synthetic images that were generated under various lighting conditions.
 
\subsection{Real images}
\begin{figure}[!tbtp]
\captionsetup[subfigure]{labelformat=empty}
\centering
\subfloat[]{\includegraphics[width=0.12\textwidth]{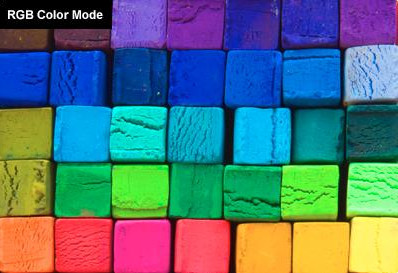}}
\hspace{0.2cm}
\subfloat[]{\includegraphics[width=0.12\textwidth]{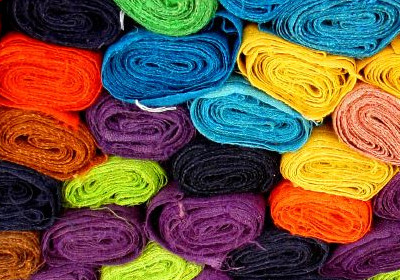}}
\hspace{0.2cm}
\subfloat[]{\includegraphics[width=0.12\textwidth]{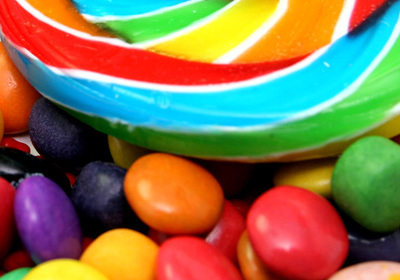}}\\
\vspace*{-\baselineskip} 
\subfloat[]{\includegraphics[width=0.12\textwidth]{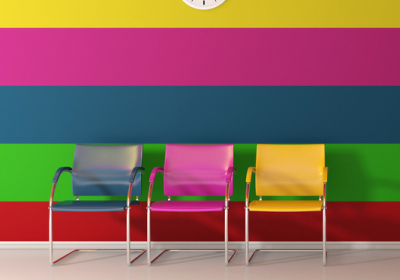}}
\hspace{0.2cm}
\subfloat[]{\includegraphics[width=0.12\textwidth]{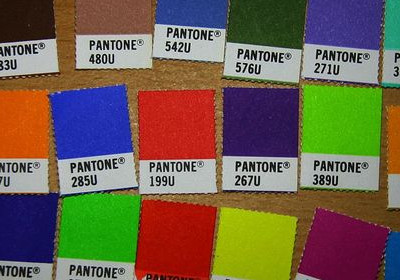}}
\hspace{0.2cm}
\subfloat[]{\includegraphics[width=0.12\textwidth]{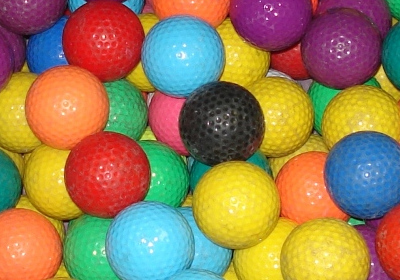}}\\
\vspace*{-\baselineskip} 
\subfloat[]{\includegraphics[width=0.12\textwidth]{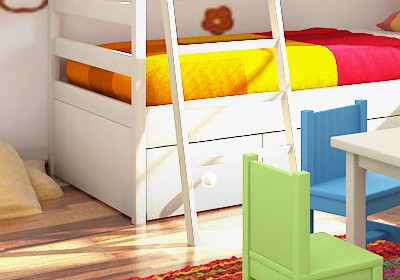}}
\hspace{0.2cm}
\subfloat[]{\includegraphics[width=0.12\textwidth]{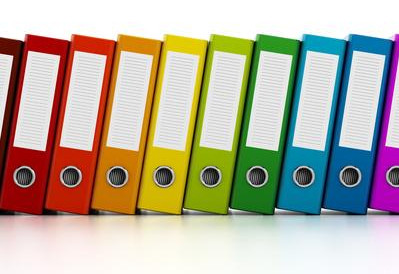}}
\hspace{0.2cm}
\subfloat[]{\includegraphics[width=0.12\textwidth]{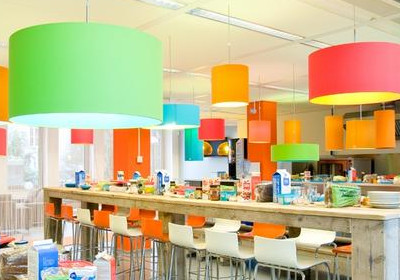}}\\
\vspace*{-\baselineskip} 
\caption{Real image samples collected from the web.}
\vspace*{-\baselineskip} 
\label{fig:real_samples}
\end{figure}

A total of 60 real images are collected from the web representing different materials, object shapes and lighting conditions. Unlike the synthetic images, the real images only contain a subset of the 12 colors. Therefore we selected them  so that all colors are reasonably distributed. Figure \ref{fig:real_samples} illustrates some of the images used in the experiments.

\section{Experiments}
We evaluated our samples both with and without normalization (pixel-wise normalization of each channel). In the latter method pixel-wise normalization was used to reduce the effect of illumination changes. The normalization took place on the original RGB images and was computed by dividing each pixel value in each channel by the sum of values in all channels.

In the following subsection we use the following abbreviations for each color category: blue-green (bg), blue (b), green (g), green-yellow(gy), orange (o), orange-red (or), red (r), red-violet (rv), violet-blue (vb), violet (v), yellow-orange (yo), and yellow (y). 

\subsection{Color spaces in the simulated environment}
\label{colorspace_in_sim}
\subsubsection{Backprojection}
Performing backprojection (BP) on the original images without pixel-wise normalization, the scores in the majority of the color spaces are fairly low. This is due to the high degree of illumination changes in the images. The only color spaces that perform well are the photometric color invariants (\textit{C1C2C3}, \textit{NOPP} and \textit{rg}) which describe color configuration of image discounting the effect of shadow or  highlights. 

To compensate for the changes in illumination, the spaces \textit{NOPP} and \textit{rg} normalize the pixel values of each channel by dividing by the sum of pixel values in all channels. The \textit{C1C2C3} color space achieves the illumination invariant behavior by calculating each channel value by $\arctan(a,\max(b,c))$  where $ a, b, c = {R,G,B} $ and $b \neq c \neq a$. 

After pixel-wise normalization, the results are changed significantly. The highlight is that the scores of the top three color spaces before normalization deteriorate due to loss of information as a result of double normalization. On the contrary, all other methods perform dramatically better due to reducing the effect of illumination changes.

\begin{figure}[!tbtp]
\vspace*{-\baselineskip}
\centering
\subfloat[Recall]{\includegraphics[width=0.7\columnwidth]{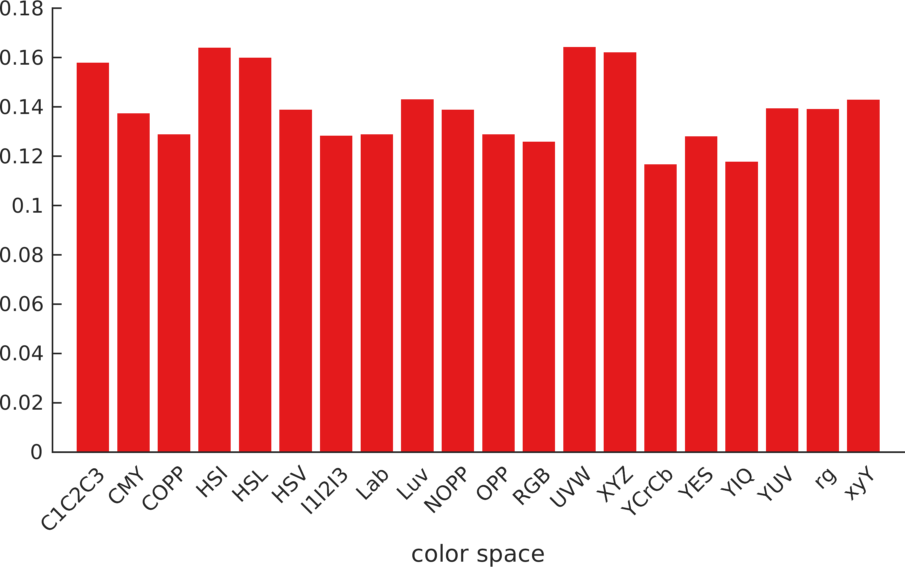}}\\
\subfloat[Precision]{\includegraphics[width=0.7\columnwidth]{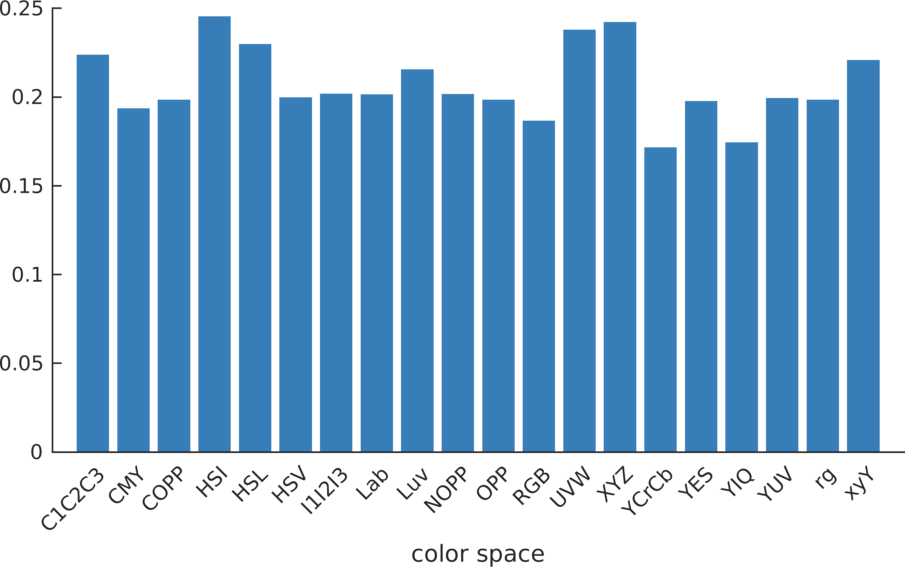}}\\
\subfloat[FMeasure]{\includegraphics[width=0.7\columnwidth]{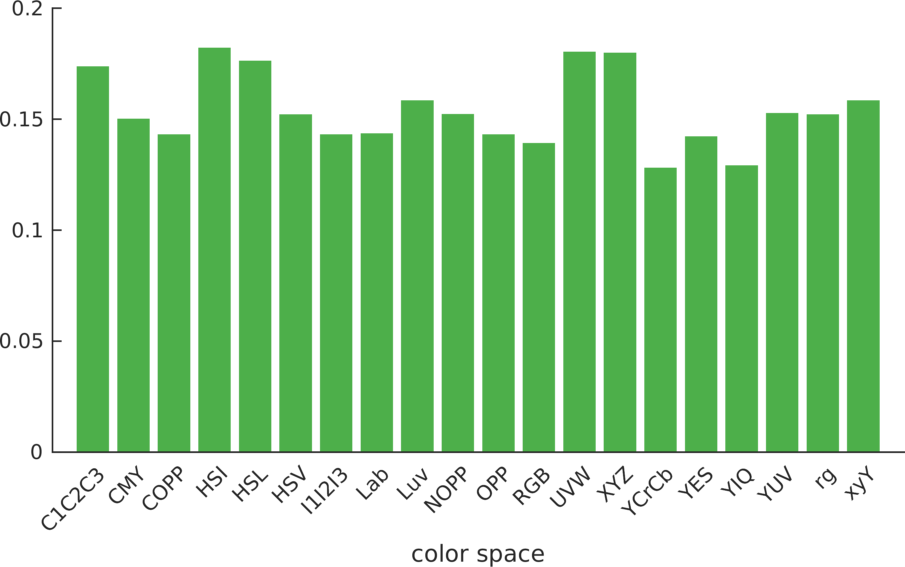}}\\
\caption{The overall results of backprojecion on normalized synthetic images.}
\label{fig:sim_overall_norm}
\end{figure}

Figure \ref{fig:sim_overall_norm} shows the scores after normalization. The best performance is achieved in color spaces \textit{HSI$^\prime$}, \textit{UVW$^\prime$} and \textit{XYZ$^\prime$}. The prime sign ($^\prime$) indicates the color space is applied after pixel-wise normalization of the input image. In addition, despite the drop in the performance of \textit{C1C2C3}, it still remains one of the top 5 color spaces.
 
\begin{figure}[!tbtp]
\vspace*{-\baselineskip}
\centering
\includegraphics[width=0.8\columnwidth]{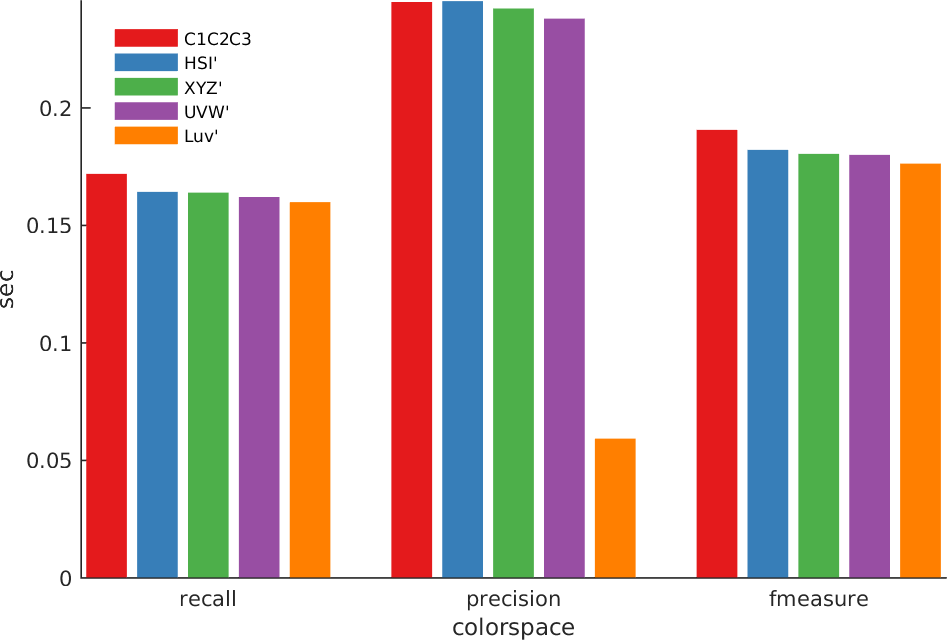}
\caption{The overall results of backprojecion in both original and normalized synthetic images. The color spaces noted by "$^\prime$" are the ones after normalization.}
\label{fig:sim_overall_total}
\end{figure}

After combining the results of both experiments, as shown in Figure \ref{fig:sim_overall_total}, BP performs best in the \textit{C1C2C3} color space followed by \textit{UVW$^\prime$} and \textit{HSI$^\prime$}.

Although BP has the best performance on average in the spaces in Figure \ref{fig:sim_overall_total}, it does not necessarily perform best in detecting each color group using those color spaces. In fact, using the \textit{C1C2C3} color space, the BP algorithm does not have the best hit rate in any of the color categories.

To highlight the performance of BP using different color spaces to detect each color group, in Table \ref{table_sim_colors} we list the top 3 color spaces followed by the highest score for the corresponding color group. 

\begin{table*}[!hbtp]
\caption{The BP results for the best performance in detecting each color in the synthetic images. The color space(s) in \textbf{bold} is (are) the best color space(s).} 
\vspace*{-\baselineskip} 
\label{table_sim_colors}
\begin{center}
\resizebox{\textwidth}{!}{
\begin{tabular}{|*{1}{c||}*{3}{c|}*{1}{c||}*{3}{c|}*{1}{c||}*{4}{c|}}
\hline

\multicolumn{1}{|c||}{\multirow{2}{*}{\textit{Color}}}&
\multicolumn{4}{c||}{Recall}&\multicolumn{4}{c||}{Precision}& \multicolumn{4}{c|}{FMeasure}\\ \cline{2-13}

\multicolumn{1}{|c||}{}& \textit{\#1} & \textit{\#2} & \textit{\#3} & \textit{Best Score}&
						\textit{\#1} & \textit{\#2} & \textit{\#3} & \textit{Best Score}&
						\textit{\#1} & \textit{\#2} & \textit{\#3} & \textit{Best Score}\\
\hline
\rowcolor{bg!20}
\textit{\textbf{bg}}  & \textbf{YCrCb$^\prime$} & \textbf{YUV$^\prime$} & $rg$  & $0.272878$ &
				        \textbf{YCrCb$^\prime$} & \textbf{YUV$^\prime$} & \textbf{YIQ$^\prime$} & $0.333333$ &
						\textbf{YCrCb$^\prime$} & \textbf{YUV$^\prime$} & $rg$ & $0.295989$\\
\hline
\rowcolor{b!20}
\textit{\textbf{b}}  & \textbf{HSI$^\prime$} & $C1C2C3$ & $XYZ^\prime$ & $0.272862$ &
					   \textbf{C1C2C3} & \textbf{NOPP} & \textbf{rg} & $0.333333$ &
					   \textbf{C1C2C3} & \textbf{NOPP} & \textbf{rg} &  $0.29595$\\
\hline
\rowcolor{g!20}
\textit{\textbf{g}}  & \textbf{HSI$^\prime$} & $C1C2C3$ & $rg$ & $0.272856$ &
					   \textbf{C1C2C3} & \textbf{NOPP} & \textbf{rg} & $0.333333$ &
					   \textbf{C1C2C3} & \textbf{NOPP} & \textbf{rg} & $0.29595$\\
\hline
\rowcolor{gy!20}
\textit{\textbf{gy}}  & \textbf{XYZ$^\prime$} & $xyY^\prime$ & $rg$ & $0.029288$ &
						\textbf{XYZ$^\prime$} & $xyY^\prime$ & $HSI^\prime$  & $0.176868$ &
 						\textbf{XYZ$^\prime$} & $xyY^\prime$ & $HSI^\prime$ & $0.29595$\\
\hline
\rowcolor{o!20}
\textit{\textbf{o}} & \textbf{HSI$^\prime$} & $HSL^\prime$ & $XYZ^\prime$ & $0.234302$ &
					  \textbf{HSL$^\prime$} & $XYZ^\prime$ & $HSI^\prime$ & $0.318852$ &
					  \textbf{HSI$^\prime$} & $HSL^\prime$ & $XYZ^\prime$ & $0.26348$ \\
\hline
\rowcolor{or!20}
\textit{\textbf{or}} & \textbf{Luv$^\prime$} & $XYZ^\prime$ & $HSI^\prime$ & $0.058072$ &
					   \textbf{Luv$^\prime$} & $XYZ^\prime$ & $UVW^\prime$ & $0.23439$ &
			           \textbf{Luv$^\prime$} & $XYZ^\prime$ & $HSI^\prime$ & $0.08608$\\
\hline
\rowcolor{r!20}
\textit{\textbf{r}} & \textbf{HSI$^\prime$} & $XYZ^\prime$ & $C1C2C3$ & $ 0.272925$ &
					  \textbf{HSI$^\prime$} & \textbf{XYZ$^\prime$} &\textbf{C1C2C3} & $ 0.333333$ &
				      \textbf{HSI$^\prime$} & \textbf{HSL$^\prime$} &\textbf{C1C2C3} & $ 0.29595$\\
\hline
\rowcolor{rv!20}
\textit{\textbf{rv}}& \textbf{UVW$^\prime$} & $C1C2C3$ & $HSI^\prime$  & $0.271568$ &
					  \textbf{C1C2C3} &  \textbf{xyY$^\prime$} & $HSV^\prime$ & $0.326746$ &
					  \textbf{UVW$^\prime$} & $C1C2C3$ & $HSI^\prime$ & $0.288187$\\
\hline
\rowcolor{vb!20}
\textit{\textbf{vb}} & \textbf{UVW$^\prime$} &  $C1C2C3$ &  $xyY^\prime$ & $0.24965$ &
					   \textbf{UVW$^\prime$} &  $C1C2C3$ &  $Lab^\prime$  & $0.32683$ &
					   \textbf{UVW$^\prime$} &  $C1C2C3$ &  $HSI^\prime$  & $0.277403$ \\
\hline
\rowcolor{v!20}
\textit{\textbf{v}} & \textbf{UVW$^\prime$} &  $HSI^\prime$  &  $C1C2C3$ & $0.019721$ &
					  \textbf{C1C2C3} &  $HSI^\prime$  &  $UVW^\prime$  & $0.3015$ &
					  \textbf{C1C2C3} &  $HSI^\prime$  &  $UVW^\prime$  & $0.208202$\\
\hline
\rowcolor{yo!20}
\textit{\textbf{yo}} & \textbf{XYZ$^\prime$} & $UVW^\prime$  & $HSL^\prime$ & $0.011852$&
 					   \textbf{XYZ$^\prime$} & $UVW^\prime$  & $HSL^\prime$  & $0.110309$ &
 			           \textbf{XYZ$^\prime$} & $UVW^\prime$  & $HSL^\prime$ & $0.030211$ \\
  \hline
  \rowcolor{y!20}
\textit{\textbf{y}} & \textbf{XYZ$^\prime$}  & $UVW^\prime$   & $rg$ & $0.283732$ &
             	      \textbf{NOPP} & $rg$   & $YCrCb^\prime$ & $0.321229$ &
 			          \textbf{rg}   & $NOPP$ & $COPP^\prime$  & $0.29336$ \\
\hline
\end{tabular}
}
\end{center}
\end{table*}

The best hit rate overall (recall) is achieved using \textit{HSI$^\prime$}, \textit{UVW$^\prime$} and \textit{XYZ$^\prime$}. \textit{HSI$^\prime$} is most suitable when the color is concentrated only in a single channel. On the other hand, \textit{UVW$^\prime$} and \textit{XYZ$^\prime$} are better for colors containing violet (red-blue) and yellow (green-red) respectively.

Using the \textit{C1C2C3} color space, the best precision and FMeasure are obtained overall. As for the recall, in four cases the second best performance is achieved using this color space and in  the cases where colors yellow or orange are present, using \textit{C1C2C3} is not advantageous. 

\subsubsection{Clustering}
We performed the EM clustering in both original and normalized images. The maximum number of clusters was set to the maximum number of colors present in the image (in this case 12) and the silhouette score was measured for each color space.

\begin{figure}[!tbtp]
\centering
\subfloat[Original image]{\includegraphics[width=0.7\columnwidth]{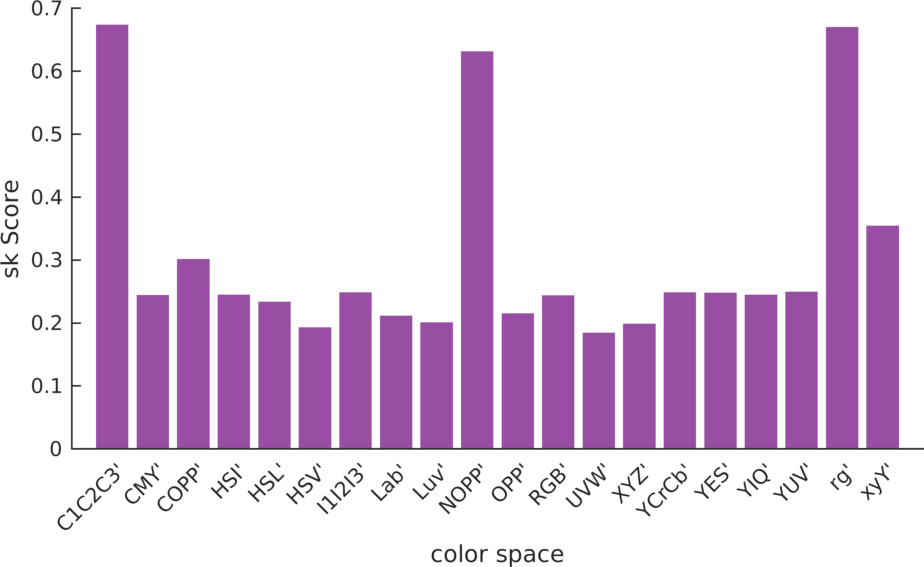}}\\
\subfloat[After normalization]{\includegraphics[width=0.7\columnwidth]{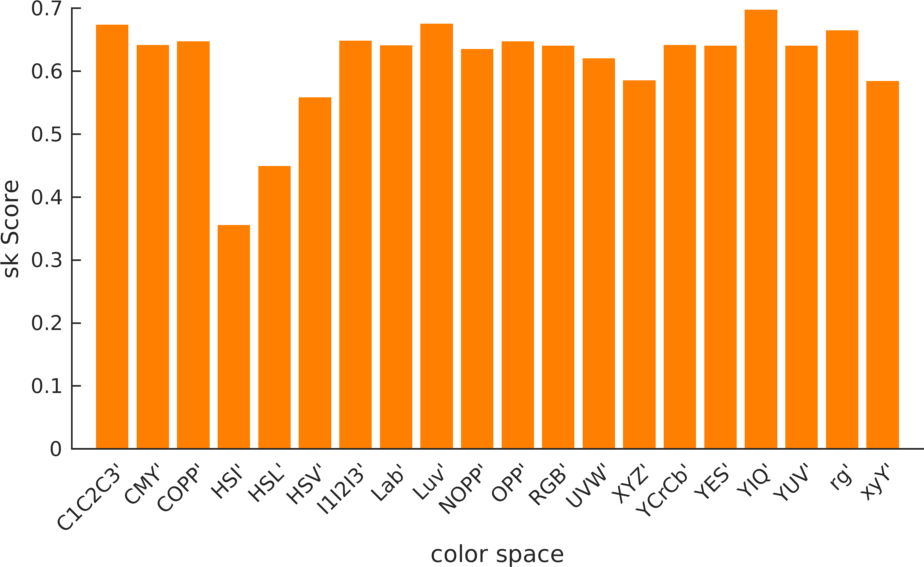}}\\
\caption{The silhouette scores of clustering using different color spaces.}
\label{fig:sim_EM}
\end{figure}

Figure \ref{fig:sim_EM} shows the results of the clustering. As  expected, after normalization, the clustering is improved. Considering the results in both scenarios, the top three color spaces are \textit{YIQ$^\prime$} (0.6977), \textit{C1C2C3} (0.6741) and \textit{rg} (0.6701). 

\subsection{Color spaces in the real images}
\subsubsection{Backprojection}

We followed the same procedure as for the simulated images and ran the BP algorithm in both original and pixel-wise normalized real images. The templates for BP are generated using a color checker. 

\begin{figure}[!tbtp]
\centering
\includegraphics[width=0.8\columnwidth]{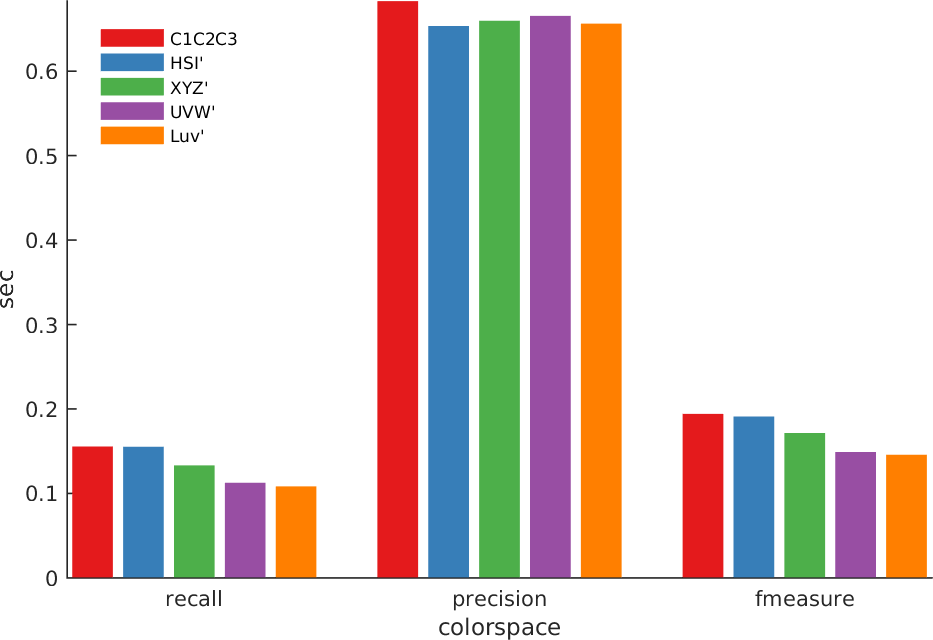}\\
\caption{The overall results of backprojection in both original and normalized real images. The color spaces noted by "$^\prime$" are the ones with normalization.}
\label{fig:real_overall_total}
\end{figure}

Once again without normalization the performance using the majority of color spaces is poor. After normalization, however, the best performance is achieved using color spaces \textit{C1C2C3$^\prime$} and \textit{UVW$^\prime$} followed by \textit{XYZ$^\prime$} as shown in Figure \ref{fig:real_overall_total}. It should be noted that the performance in \textit{C1C2C3} is also slightly improved after normalizing the image.

\begin{figure}[!tbtp]
\vspace*{-\baselineskip}
\centering
\subfloat[Recall]{\includegraphics[width=0.7\columnwidth]{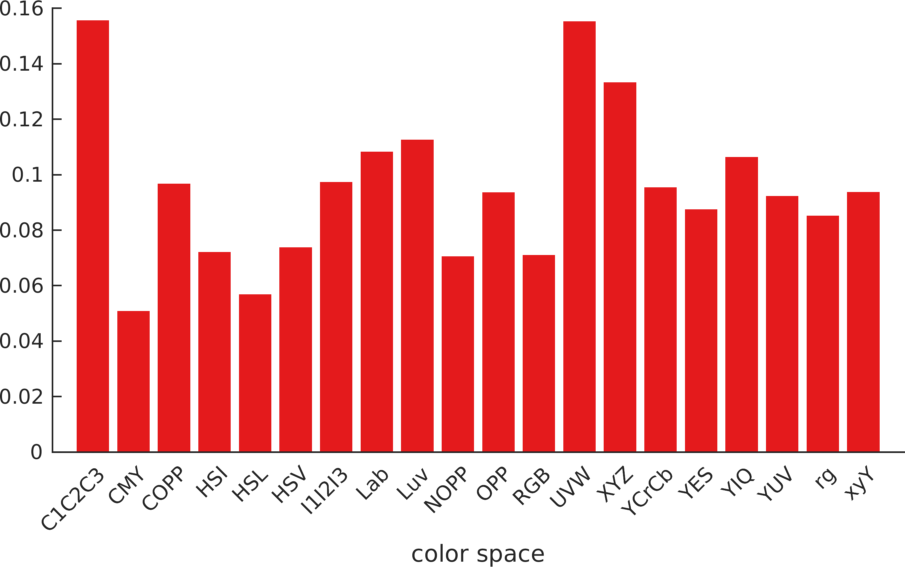}}\\
\subfloat[Precision]{\includegraphics[width=0.7\columnwidth]{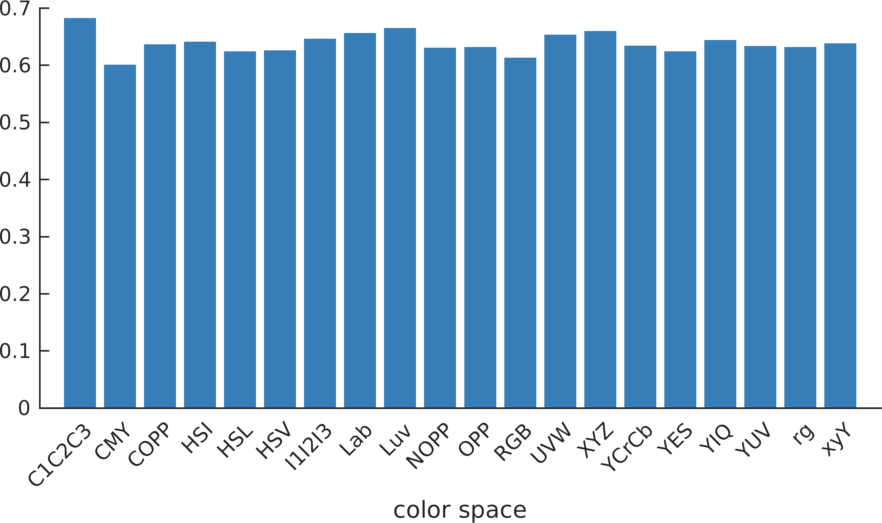}}\\
\subfloat[FMeasure]{\includegraphics[width=0.7\columnwidth]{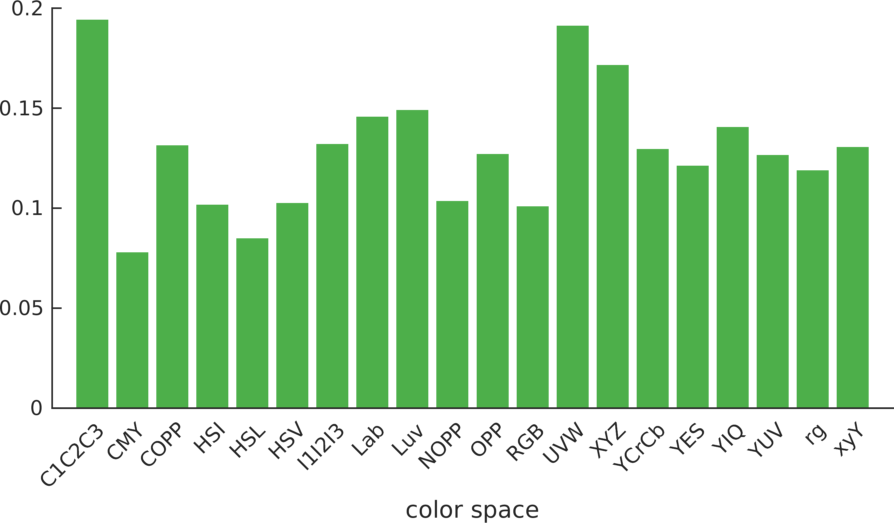}}\\
\caption{The overall results of backprojecion on normalized real images.}
\label{fig:real_overall_norm}
\end{figure}

The performance of BP in detecting different color groups in different color spaces is reflected in Table \ref{table_real_colors}.

\begin{table*}[!hbtp]
\caption{The BP results for the best performance in detecting each color in real images. The color space in \textbf{bold} is the best color space.} 
\vspace*{-\baselineskip} 
\label{table_real_colors}
\begin{center}
\resizebox{\textwidth}{!}{
\begin{tabular}{|*{1}{c||}*{3}{c|}*{1}{c||}*{3}{c|}*{1}{c||}*{4}{c|}}
\hline
\multicolumn{1}{|c||}{\multirow{2}{*}{\textit{Color}}}&
\multicolumn{4}{c||}{Recall}&\multicolumn{4}{c||}{Precision}& \multicolumn{4}{c|}{FMeasure}\\ \cline{2-13}

\multicolumn{1}{|c||}{}& \textit{\#1} & \textit{\#2} & \textit{\#3} & \textit{Best Score}&
						\textit{\#1} & \textit{\#2} & \textit{\#3} & \textit{Best Score}&
						\textit{\#1} & \textit{\#2} & \textit{\#3} & \textit{Best Score}\\
\hline
\rowcolor{bg!20}
\textit{\textbf{bg}}  & \textbf{UVW$^\prime$} & $C1C2C3$ & $YIQ^\prime$  & $0.218664$ &
				        \textbf{UVW$^\prime$} & $YIQ^\prime$ & $C1C2C3$ & $0.676459$ &
						\textbf{UVW$^\prime$} & $C1C2C3$ & $YIQ^\prime$ & $0.262503$\\
\hline
\rowcolor{b!20}
\textit{\textbf{b}}  & \textbf{C1C2C3} & $UVW^\prime$ & $YIQ^\prime$ & $0.034735$ &
					   \textbf{C1C2C3} & $Luv^\prime$ & $UVW^\prime$ & $0.672606$ &
					   \textbf{C1C2C3} & $UVW^\prime$ & $rg$ &  $0.055433$\\
\hline
\rowcolor{g!20}
\textit{\textbf{g}}  & \textbf{Lab$^\prime$} & $UVW^\prime$ & $C1C2C3$ & $0.039952$ &
					   \textbf{C1C2C3} & $HSI^\prime$ & $Luv^\prime$ & $0.620502$ &
					   \textbf{Lab$^\prime$} & $C1C2C3^\prime$ & $UVW^\prime$ & $0.060729$\\
\hline
\rowcolor{gy!20}
\textit{\textbf{gy}}  & \textbf{UVW$^\prime$} & $XYZ^\prime$ & $COPP^\prime$ & $0.186728$ &
						\textbf{UVW$^\prime$} & $rg$ & $Lab^\prime$  & $0.714204$ &
 						\textbf{UVW$^\prime$} & $XYZ^\prime$ & $COPP^\prime$ & $0.22808$\\
\hline
\rowcolor{o!20}
\textit{\textbf{o}} & \textbf{C1C2C3} & $UVW^\prime$ & $XYZ^\prime$ & $0.270723$ &
					  \textbf{UVW$^\prime$} & $Lab^\prime$ & $Luv^\prime$ & $0.681821$ &
					  \textbf{C1C2C3} & $UVW^\prime$ & $XYZ^\prime$ & $0.296339$ \\
\hline
\rowcolor{or!20}
\textit{\textbf{or}} & \textbf{C1C2C3} & $UVW^\prime$ & $XYZ^\prime$ & $0.304511$ &
					   \textbf{NOPP$^\prime$} & $HSL^\prime$ & $C1C2C3$ & $0.705755$ &
			           \textbf{C1C2C3} & $UVW^\prime$ & $XYZ^\prime$  & $0.324479$\\
\hline
\rowcolor{r!20}
\textit{\textbf{r}} & \textbf{Lab$^\prime$} & $C1C2C3$ & $UVW^\prime$  & $0.200768$ &
					  \textbf{C1C2C3} & $UVW^\prime$ & $Lab^\prime$ & $0.805447$ &
				      \textbf{C1C2C3} & $Lab^\prime$ & $UVW^\prime$ & $0.263881$\\
\hline
\rowcolor{rv!20}
\textit{\textbf{rv}}& \textbf{Luv$^\prime$} & $C1C2C3$ & $Lab^\prime$ & $0.130811$ &
					  \textbf{YIQ$^\prime$} & $UVW^\prime$ & $C1C2C3^\prime$ & $0.737866$ &
					  \textbf{Luv$^\prime$} & $C1C2C3$ & $UVW^\prime$ & $0.161168$\\
\hline
\rowcolor{vb!20}
\textit{\textbf{vb}} & \textbf{YCrCb$^\prime$} &  $YUV^\prime$ &  $Lab^\prime$ & $0.186816$ &
					   \textbf{Luv$^\prime$} &  $I1I2I3^\prime$ &  $Lab^\prime$  & $0.842699$ &
					   \textbf{YCrCb$^\prime$} &  $YUV^\prime$ &  $Lab^\prime$  & $0.234721$ \\
\hline
\rowcolor{v!20}
\textit{\textbf{v}} & \textbf{XYZ$^\prime$} &  $COPP$  &  $C1C2C3$ & $0.152557$ &
					  \textbf{XYZ$^\prime$} &  $YIQ$  &  $C1C2C3$  & $0.659558$ &
					  \textbf{COPP} &  $XYZ^\prime$  &  $C1C2C3$  & $0.196943$\\
\hline
\rowcolor{yo!20}
\textit{\textbf{yo}} & \textbf{XYZ$^\prime$} & $C1C2C3$  & $Luv^\prime$ & $0.17477$&
 					   \textbf{C1C2C3} & $XYZ^\prime$  & $CMY^\prime$  & $0.633117$ &
 			           \textbf{C1C2C3} & $UVW^\prime$  & $Luv^\prime$ & $0.22849$ \\
  \hline
  \rowcolor{y!20}
\textit{\textbf{y}} & \textbf{UVW$^\prime$}  & $XYZ^\prime$   & $YIQ^\prime$ & $0.296691$ &
             	      \textbf{YIQ$^\prime$} & $rg$   & $HSI^\prime$ & $0.79604$ &
 			          \textbf{UVW$^\prime$}   & $XYZ^\prime$ & $C1C2C3$  & $0.341509$ \\
\hline

\end{tabular}
}
\end{center}
\end{table*}

The best performance results from the \textit{C1C2C3} color space. The only cases in which the performance was poor was in color groups \textit{y} and \textit{gy} similar to  simulated images.
 
The runner-up color spaces are \textit{UVW$^\prime$} and \textit{XYZ$^\prime$}. The performance was more or less  similar in both synthetic and real images. However, there are some exceptions. For instance, in the case of color \textit{v}, the best performance belongs to \textit{UVW$^\prime$} in synthetic images where \textit{XYZ$^\prime$} is not even in top three spaces. In contrast, using real images opposite results were observed. Here, the best space is \textit{XYZ$^\prime$} whereas \textit{UVW$^\prime$} is not in the top three.                                                                                                                                                                                                                                                                                                      

\subsubsection{Clustering}

\begin{figure}[!tbtp]
\centering
\subfloat[Original image]{\includegraphics[width=0.7\columnwidth]{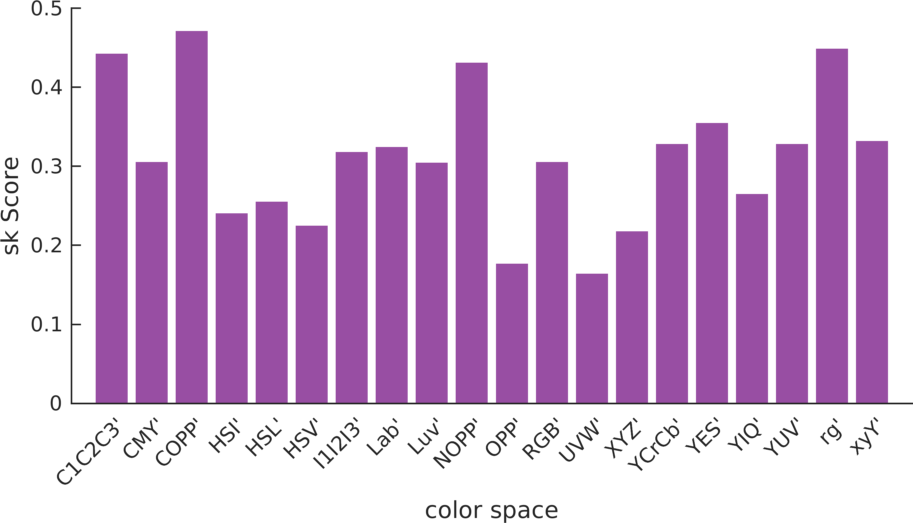}}\\
\subfloat[After normalization]{\includegraphics[width=0.7\columnwidth]{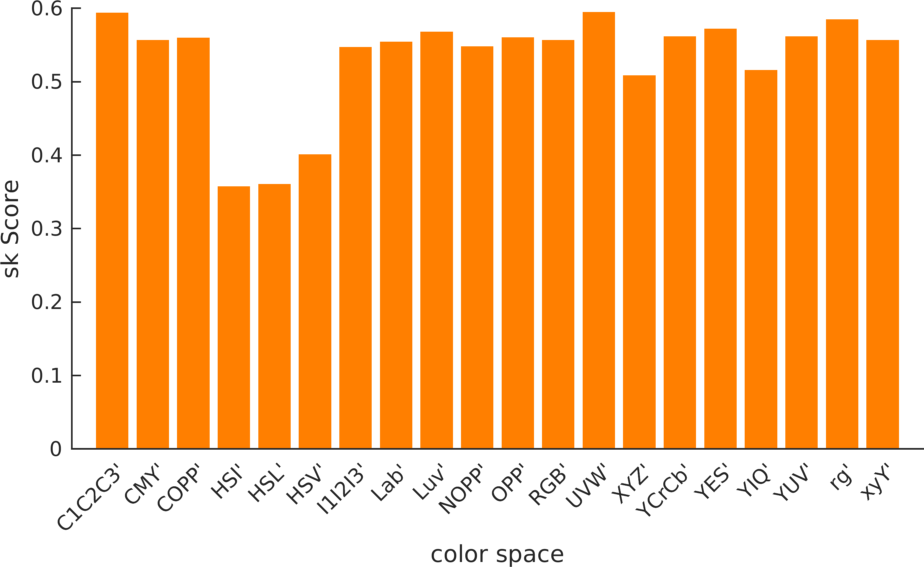}}\\
\caption{The silhouette scores of clustering using different color spaces.}

\label{fig:real_EM}
\end{figure}

The clustering was done on both original and normalized images. The maximum number of clusters was set to the maximum number of colors present in the image ranging from 4-12 colors maximum depending on the image. The silhouette scores are measured and reflected in Figure \ref{fig:real_EM}.

Similar to simulated images, normalization improves the overall results. The best performance was achieved using the color spaces \textit{COPP}, \textit{rg} and \textit{C1C2C3} in original images and \textit{UVW$^\prime$}, \textit{C1C2C3$^\prime$} and \textit{rg$^\prime$} after normalization. 

\subsection{Active visual search}
In this section we put our findings into practice and evaluate the effect of color space choice on the performance of a mobile robot searching for an object. To perform search we used the same greedy algorithm introduced in \cite{rasouli2016sensor} with the difference of omitting the bottom-up saliency to eliminate any bias beside the color values.

\begin{figure}[!tbtp]
\centering
\includegraphics[width=0.8\columnwidth]{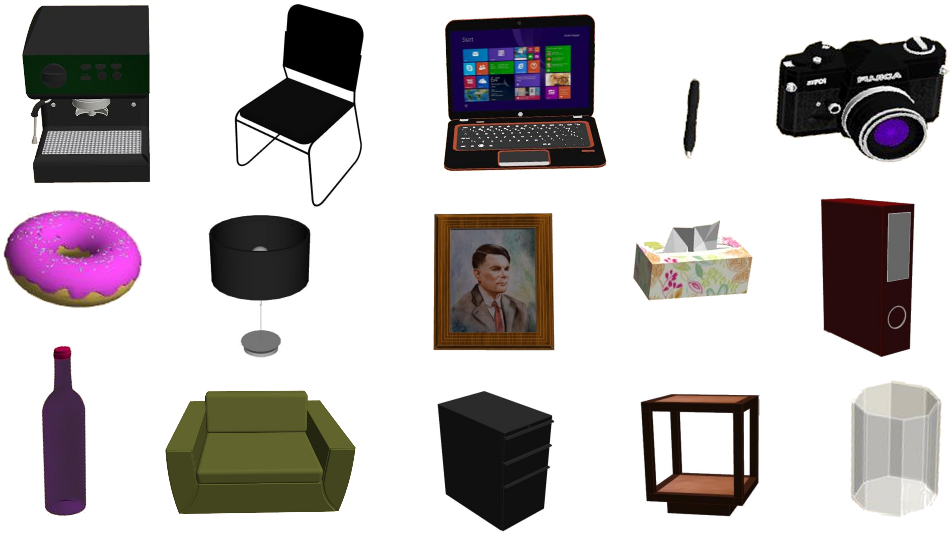}\\
\caption{Object samples used in the experiment.}
\label{fig:objects_background}
\end{figure}

\begin{figure}[!tbtp]
\centering
\includegraphics[width=\columnwidth]{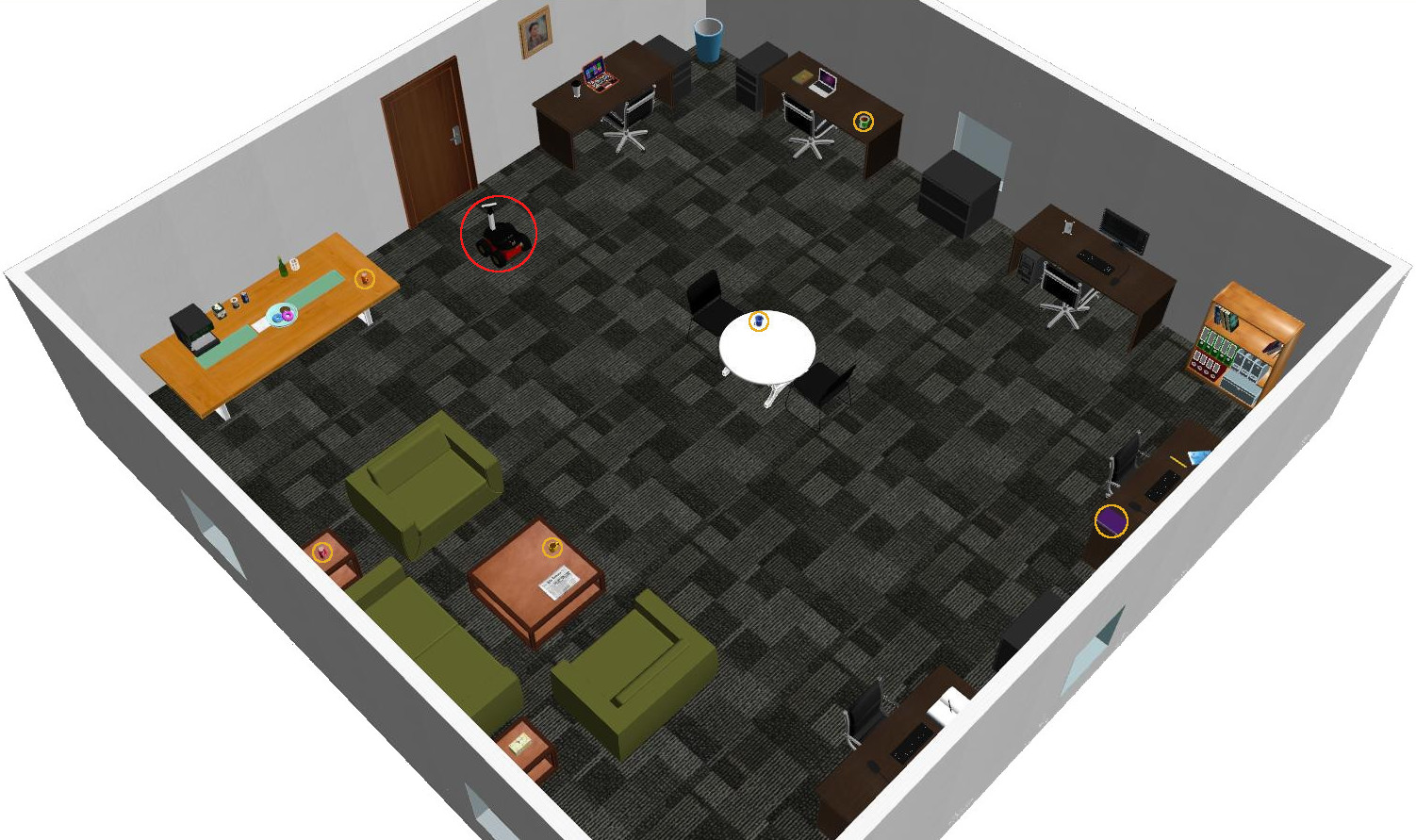}\\
\caption{The simulated office environment used for the experiments. The targets and the robot are highlighted by yellow and red circles respectively.}
\label{fig:office_env}
\end{figure}

% maybe add the full dataset link here
The experiments were conducted in Gazebo simulation environment. For this purpose we generated a large number of objects\footnote{The dataset is available at \href{http://data.nvision2.eecs.yorku.ca/3DGEMS}{http://data.nvision2.eecs.yorku.ca/3DGEMS/}} (see Figure \ref{fig:objects_background}) to create a typical office environment (see Figure \ref{fig:office_env}). The search robot is a simulated Pioneer 3 platform equipped with a Zed camera for visual processing and Hokuyo Lidar for mapping. In addition, communications and navigation are done using ROS nav package.

We used 6 target objects with various colors (see Figure \ref{fig:objects}) placed randomly in the scene. In each iteration the locations of objects were rotated, so that objects were equally represented in the environment. In each configuration, the robot was placed in four different locations to begin the search. 

\begin{figure}[!tbtp]
\centering
\includegraphics[width=0.6\columnwidth]{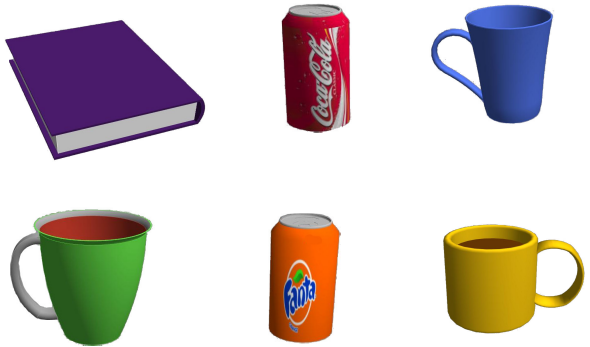}\\
\caption{Target objects used in the experiments.}
\label{fig:objects}
\end{figure}

As for the color spaces, we used the top 6 performing spaces found in Section \ref{colorspace_in_sim}, namely \textit{C1C2C3},  \textit{UVW$^\prime$}, \textit{XYZ$^\prime$}, \textit{YCrCb$^\prime$}, \textit{Luv$^\prime$} and  \textit{HSI$^\prime$}. In total over 1000 experiments were conducted.

Table \ref{table_colors_search} lists the results of experiments for each object. The outcomes are consistent with the results in Table \ref{table_sim_colors}. Using the top color spaces resulted in lower search time. There are, however, a few exceptions that show the importance of discriminability. For instance, \textit{UVW$^\prime$} is placed in top 2 position for detecting 5 color groups but in search at best is placed in 3rd and 4th position. Given this color space's low silhouette score, the detection algorithm can be distracted by the other colors in the environment, therefore, the efficiency of the color space is reduced. In contrast, using \textit{C1C2C3} better performance is achieved. This is consistent with the high silhouette score which means this space is more robust against distraction.
  
\begin{table*}[!hbtp]
\caption{The results of the search experiments for each object}
\vspace*{-\baselineskip}  
\label{table_colors_search}
\begin{center}
\resizebox{0.70\textwidth}{!}
{
\begin{tabular}{|*{1}{c||}*{6}{c||}*{1}{c|}}
\hline
Object & \textit{\#1} & \textit{\#2} & \textit{\#3} & \textit{\#4} & \textit{\#5} & \textit{\#6} & \textit{Best time(sec)}\\
\hline
\rowcolor{b!20}
\textit{\textbf{blue cup}}& \textbf{XYZ$^\prime$} & $C1C2C3$ & $HSI^\prime$&
			            $UVW^\prime$ & $YCrCb^\prime$ & $Luv^\prime$& $340.83$\\
			            \hline
\rowcolor{y!20}
\textit{\textbf{yellow cup}}& \textbf{C1C2C3} & $HSI^\prime$ & $UVW^\prime$&
			            $XYZ^\prime$ & $Luv^\prime$ & $YCrCb^\prime$& $345.83$\\
			            \hline
\rowcolor{g!20}
\textit{\textbf{green cup}}& \textbf{HSI$^\prime$} & $C1C2C3$ & $XYZ^\prime$&
			            $UVW^\prime$ & $Luv^\prime$ & $YCrCb^\prime$& $304.17$\\
			            \hline
\rowcolor{v!20}
\textit{\textbf{violet book}}& \textbf{XYZ$^\prime$} & $C1C2C3$ & $HSI^\prime$&
			            $Luv^\prime$ & $UVW\prime$ & $YCrCb^\prime$& $354.17$\\
			            \hline
\rowcolor{r!20}
\textit{\textbf{red can}}& \textbf{HSI$^\prime$} & $C1C2C3$ & $UVW^\prime$ &
			            $Luv^\prime$ & $YCrCb\prime$ & $XYZ^\prime$& $261.67$\\
			            \hline
\rowcolor{o!20}
\textit{\textbf{orange can}}& \textbf{C1C2C3} & $XYZ^\prime$ & $Luv^\prime$ &
			            $UVW^\prime$ & $YCrCb\prime$ & $HSI^\prime$& $317.50$\\
			            \hline
\end{tabular}
}
\end{center}
\end{table*}

\begin{figure}[!tbtp]
\centering
\includegraphics[width=0.6\columnwidth]{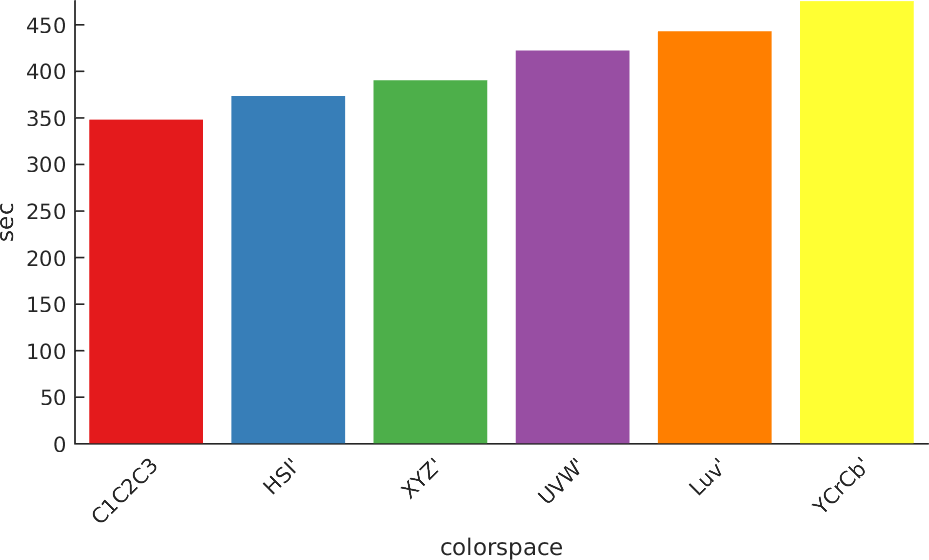}\\
\caption{The average result of the search for all objects.}
\label{fig:avg_search_result}
\end{figure}

Figure \ref{fig:avg_search_result} shows the average results of the search for all objects. Here, the best performance overall is achieved using color spaces (in descending order) \textit{C1C2C3}, \textit{HSI$^\prime$} and \textit{XYZ$^\prime$}.

\section{Conclusion}
In this paper we evaluated a large number of color spaces to measure their suitability detecting and discriminating different colored objects. Using empirical evaluations on both synthetic and real images we showed that there is no single optimal color space for detecting and discriminating all color groups. On average, however, the best performance was achieved using color spaces \textit{C1C2C3}, \textit{UVW} and \textit {XYZ}.

The color spaces were also put to test in the context of visual search in a simulated environment. A combination of high detection rate and robustness to distractors resulted in the lowest time of search using \textit{C1C2C3} and \textit {XYZ} color spaces.

We only measured the robustness to distraction by a clustering technique. It would be beneficial to measure the sensitivity of each color space to different color groups. In addition, the visual search experiments were done only in a simulated environment. In the future, we intend to perform a similar study on a practical platform to confirm our evaluation on the real images. 

\section*{Acknowledgment}
We acknowledge the financial support of the Natural Sciences and Engineering Research Council of Canada (NSERC), the NSERC Canadian Field Robotic Network (NCFRN), and the Canada Research Chairs Program through grants to JKT.
\bibliographystyle{IEEEtran}
{\small

\bibliography{ar,color}

% Generated by IEEEtran.bst, version: 1.13 (2008/09/30)
\begin{thebibliography}{10}
\providecommand{\url}[1]{#1}
\csname url@samestyle\endcsname
\providecommand{\newblock}{\relax}
\providecommand{\bibinfo}[2]{#2}
\providecommand{\BIBentrySTDinterwordspacing}{\spaceskip=0pt\relax}
\providecommand{\BIBentryALTinterwordstretchfactor}{4}
\providecommand{\BIBentryALTinterwordspacing}{\spaceskip=\fontdimen2\font plus
\BIBentryALTinterwordstretchfactor\fontdimen3\font minus
  \fontdimen4\font\relax}
\providecommand{\BIBforeignlanguage}[2]{{%
\expandafter\ifx\csname l@#1\endcsname\relax
\typeout{** WARNING: IEEEtran.bst: No hyphenation pattern has been}%
\typeout{** loaded for the language `#1'. Using the pattern for}%
\typeout{** the default language instead.}%
\else
\language=\csname l@#1\endcsname
\fi
#2}}
\providecommand{\BIBdecl}{\relax}
\BIBdecl

\bibitem{Moroney1995luv}
N.~Moroney and D.~F. Fairchild, ``Color space selection for jpeg image
  compression,'' \emph{Journal of Electronic Imaging}, vol.~4, no.~4, pp.
  373--381, 1995.

\bibitem{Saber1996yes}
E.~Saber, A.~Tekalp, R.~Eschbach, and K.~Knox, ``Automatic image annotation
  using adaptive color classification,'' \emph{Graphical Models and Image
  Processing}, vol.~58, no.~2, pp. 115--126, March 1996.

\bibitem{Kumar2015}
A.~Kumar and S.~Malhotra, ``{Pixel-Based Skin Color Classifier : A Review},''
  \emph{International Journal of Signal Processing, Image Processing and
  Pattern Recognition}, vol.~8, no.~7, pp. 283--290, 2015.

\bibitem{Danelljan2014hsv}
M.~Danelljan, F.~S. Khan, M.~Felsberg, and J.~V. de~Weijer, ``Adaptive color
  attributes for real-time visual tracking,'' in \emph{he IEEE Conference on
  Computer Vision and Pattern Recognition}, 2014, pp. 1090--1097.

\bibitem{Liang2015opp}
P.~Liang, E.~Blasch, and H.~Ling, ``Encoding color information for visual
  tracking: algorithms and benchmark,'' \emph{IEEE Transactions on Image
  Processing}, vol.~24, no.~12, pp. 5630--5644, 2015.

\bibitem{Stokman2005}
H.~Stokman and T.~Gevers, ``{Selection and Fusion of Color Models for Feature
  Detection.pdf},'' \emph{CVPR}, 2005.

\bibitem{Ohta1980}
Y.-I. Ohta, T.~Kanade, and T.~Sakai, ``Color information for region
  segmentation,'' \emph{Computer graphics and image processing}, vol.~13,
  no.~3, pp. 222--241, 1980.

\bibitem{Meas-Yedid2004}
V.~Meas-Yedid, E.~Glory, E.~Morelon, C.~Pinset, G.~Stamon, and J.~C.
  Olivo-Marin, ``{Automatic color space selection for biological image
  segmentation},'' \emph{Proceedings - International Conference on Pattern
  Recognition}, vol.~3, pp. 514--517, 2004.

\bibitem{Dev2014}
S.~Dev, Y.~H. Lee, and S.~Winkler, ``{Systematic study of color spaces and
  components for the segmentation of sky/cloud images},'' \emph{2014 IEEE
  International Conference on Image Processing, ICIP 2014}, pp. 5102--5106,
  2014.

\bibitem{Gupta2016}
A.~Gupta and A.~Chaudhary, ``{Robust skin segmentation using color space
  switching},'' \emph{Pattern Recognition and Image Analysis}, vol.~26, no.~1,
  pp. 61--68, 2016.

\bibitem{Vezhnevets2003}
V.~Vezhnevets, ``{A Survey on Pixel-Based Skin Color Detection Techniques},''
  \emph{Cybernetics}, vol.~85, no. 0896-6273 SB - IM, pp. 85--92, 2003.

\bibitem{Benedek2007}
C.~Benedek and T.~Szir{\'{a}}nyi, ``{Study on color space selection for
  detecting cast shadows in video surveillance},'' \emph{International Journal
  of Imaging Systems and Technology}, vol.~17, no.~3, pp. 190--201, 2007.

\bibitem{VanDeSande2010}
K.~{Van De Sande}, T.~Gevers, and C.~Snoek, ``{Evaluating color descriptors for
  object and scene recognition},'' \emph{IEEE Transactions on Pattern Analysis
  and Machine Intelligence}, vol.~32, no.~9, pp. 1582--1596, 2010.

\bibitem{Scandaliaris2007}
J.~Scandaliaris, M.~Villamizar, J.~Andrade-Cetto, and A.~Sanfeliu, ``{Robust
  color contour object detection invariant to shadows},'' \emph{Proceedings of
  the Congress on pattern recognition 12th Iberoamerican conference on Progress
  in pattern recognition, image analysis and applications}, pp. 301--310, 2007.

\bibitem{Paschos2001}
G.~Paschos, ``{Perceptually uniform color spaces for color texture analysis:
  an$\backslash$nempirical evaluation},'' \emph{IEEE Transactions on Image
  Processing}, vol.~10, no.~6, pp. 932--937, 2001.

\bibitem{Porebski2007}
A.~Porebski and N.~Vandenbroucke, ``{ Iterative Feature Selection for Color
  Texture Classification Ecole d'Ingenieurs du Pas-de-Calais Departement
  Automatique Campus de la Malassise 62967 Longuenesse Cedex - France
  Laboratoire LAGIS - UMR CNRS 8146 Universite des Sciences et Techno},''
  \emph{Image Processing, 2007. ICIP 2007. IEEE International Conference on},
  pp. 509--512, 2007.

\bibitem{Song2010}
D.-L. Song, L.-H. Ge, W.-W. Qi, and M.~Chen, ``{Illumination invariant color
  model selection based on genetic algorithm in robot soccer},''
  \emph{Information Science and Engineering (ICISE), 2010 2nd International
  Conference on}, no.~3, pp. 1--4, 2010.

\bibitem{Song2014}
D.~Song, W.~Sun, Z.~Ji, G.~Hou, X.~Li, and L.~Liu, ``{Color model selection for
  underwater object recognition},'' \emph{International Conference on
  Information Science, Electronics and Electrical Engineering}, pp. 1339--1342,
  2014.

\bibitem{Duan2015}
G.~Duan, F.~Duan, Y.~Xu, H.~Gong, and X.~Qu, ``{Investigation of Optimal
  Segmentation Color Space of Bayer True Color Images with Multi-Objective
  Optimization Methods},'' \emph{Journal of the Indian Society of Remote
  Sensing}, vol.~43, no.~3, pp. 487--499, 2015.

\bibitem{Rasouli2014vs}
A.~Rasouli and J.~K. Tsotsos, ``Attention in autonomous robotic visual
  search,'' in \emph{i-SAIRAS}, Montreal, June 2014.

\bibitem{Guo2000hsi}
P.~Guo and M.~R. Lyu, ``A study on color space selection for determining image
  segmentation region number,'' in \emph{the 2000 International Conference on
  Artificial Intelligence (IC-AI’2000)}, Las Vegas, 2000, pp. 1127--1132.

\bibitem{salvador2004cast}
E.~Salvador, A.~Cavallaro, and T.~Ebrahimi, ``Cast shadow segmentation using
  invariant color features,'' \emph{Computer vision and image understanding},
  vol.~95, no.~2, pp. 238--259, 2004.

\bibitem{Tkalcic2013yuv}
M.~Tkalcic and J.~F. Tasic, ``Colour spaces: perceptual, historical and
  applicational background,'' in \emph{Eurocon}, 2013.

\bibitem{Weeks1995hsl}
A.~R. Weeks, C.~E. Felix, and H.~R. Myler, ``Edge detection of color images
  using the hsl color space,'' in \emph{In IS\&T/SPIE's Symposium on Electronic
  Imaging: Science \& Technology}, 1995, pp. 291--301.

\bibitem{Ohta1980uvw}
Y.~I. Ohta, T.~Kanade, and T.~Sakai, ``Color information for region
  segmentation,'' \emph{Computer graphics and image processing}, vol.~13,
  no.~3, pp. 222--241, 1980.

\bibitem{Lucchese2000xyy}
L.~Lucchese and S.~K. Mitra, ``Filtering color images in the xyy color space,''
  in \emph{ICIP}, 2000, pp. 500--503.

\bibitem{Swain1991}
M.~J. Swain and D.~H. Ballard, ``Color indexing,'' \emph{Computer Vision},
  vol.~7, no.~1, pp. 11--32, 1991.

\bibitem{rousseeuw1987silhouettes}
P.~J. Rousseeuw, ``Silhouettes: a graphical aid to the interpretation and
  validation of cluster analysis,'' \emph{Journal of computational and applied
  mathematics}, vol.~20, pp. 53--65, 1987.

\bibitem{rasouli2016sensor}
A.~Rasouli and J.~K. Tsotsos, ``Sensor planning for 3d visual search with task
  constraints,'' in \emph{Computer and Robot Vision (CRV), 2016 13th Conference
  on}.\hskip 1em plus 0.5em minus 0.4em\relax IEEE, 2016, pp. 37--44.

\end{thebibliography}
}
\end{document}